\documentclass{article}

\usepackage{microtype}
\usepackage{graphicx}
\usepackage{subcaption}
\usepackage{booktabs} % for professional tables
\usepackage{enumitem}

\usepackage{hyperref}

\usepackage[accepted]{icml2026}

\usepackage{amsmath}
\usepackage{amssymb}
\usepackage{mathtools}
\usepackage{amsthm}

\usepackage[capitalize,noabbrev]{cleveref}

\theoremstyle{plain}

\theoremstyle{definition}

\theoremstyle{remark}

\usepackage[textsize=tiny]{todonotes}

\usepackage{xcolor}

\usepackage{placeins} % \FloatBarrier to force pending floats to flush

\icmltitlerunning{DragOn: A Benchmark and Dataset for Drag-Based GUI Interactions}

\begin{document}

\twocolumn[
  \icmltitle{DragOn: A Benchmark and Dataset\\ for Drag-Based GUI Interactions}



  \icmlsetsymbol{equal}{*}

  \begin{icmlauthorlist}
    \icmlauthor{Nathan Bout}{H}
    \icmlauthor{Maxime Langevin}{H}
    \icmlauthor{Ronan Riochet}{H}
  \end{icmlauthorlist}

  \icmlaffiliation{H}{H Company}
  \icmlcorrespondingauthor{Nathan Bout}{nathan.bout@hcompany.ai}
  \icmlcorrespondingauthor{Ronan Riochet}{ronan.riochet@hcompany.ai}

  \icmlkeywords{Computer-Use, VLM, ICML}

  \vskip 0.3in
]



\printAffiliationsAndNotice{}  % no special notice (required even if empty)

\begin{abstract}
GUI agents -- vision-based models that control desktops, web browsers, and mobile devices through graphical user interfaces -- promise to automate a wide range of digital tasks.
While million-scale datasets have enabled substantial progress on click-grounding, drag grounding (e.g. drag-and-drop, swipe, highlight) data remains an order of magnitude smaller and current models fall short on complex drag-based interactions.
We introduce \textit{DragOn}, a drag grounding benchmark and training dataset covering four domains: text highlighting, cell selection, element resizing and slider manipulation.
The dataset comprises 286K training screenshots and 3.5M training tasks, plus a 2{,}000-example held-out evaluation suite.
We evaluate proprietary (GPT, Claude) and open-weight (Qwen, Kimi, Holo) models, as well as a Qwen VLM fine-tuned on our training data. 
Results suggest that our dataset could improve performance of state-of-the-art models on downstream computer-use tasks.

\end{abstract}

\section{Introduction}

\begin{figure*}[t]
  \centering
  \begin{subfigure}[b]{0.495\textwidth}
    \centering
    \includegraphics[width=\linewidth]{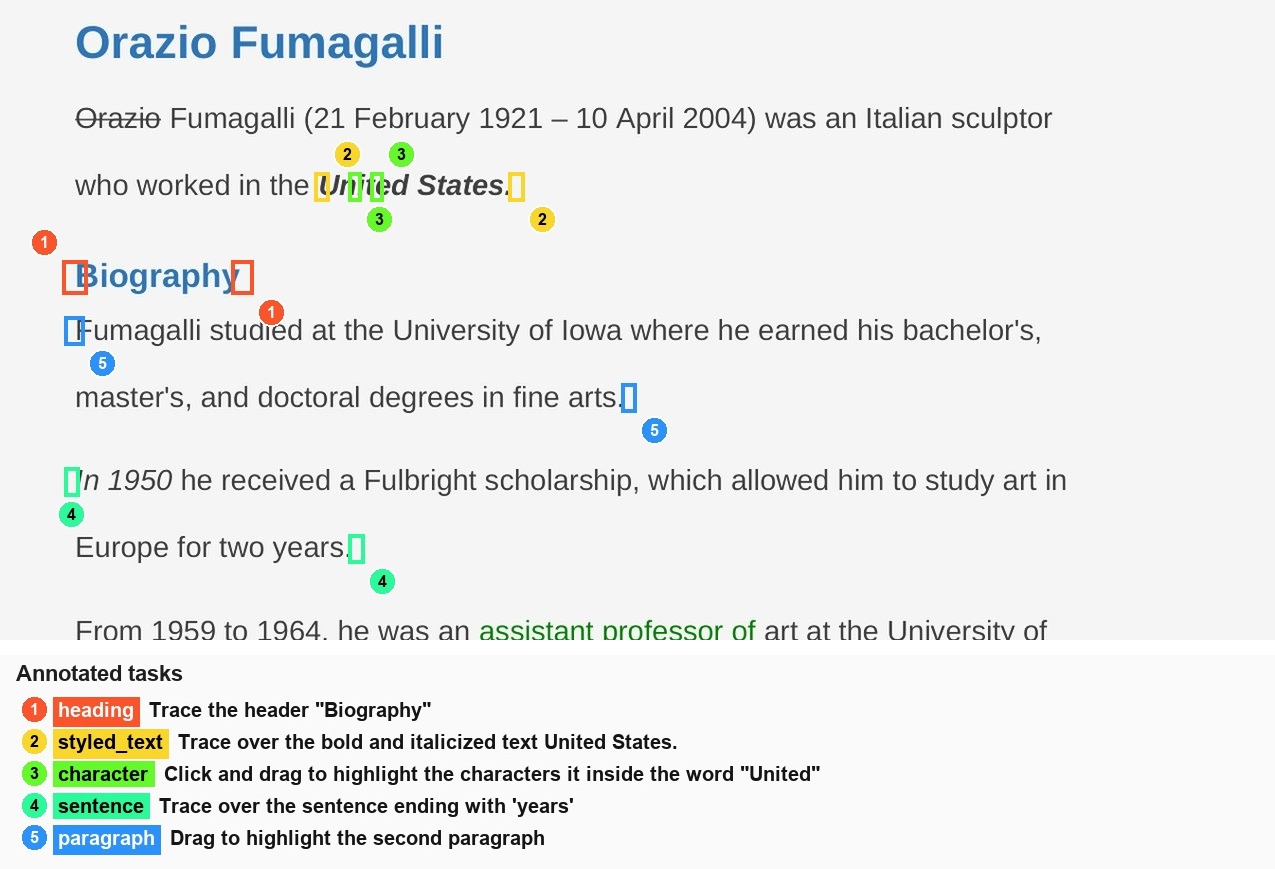}
    \caption{\textit{Text highlighting}}
    \label{fig:teaser-text}
  \end{subfigure}
  \hfill
  \begin{subfigure}[b]{0.495\textwidth}
    \centering
    \includegraphics[width=\linewidth]{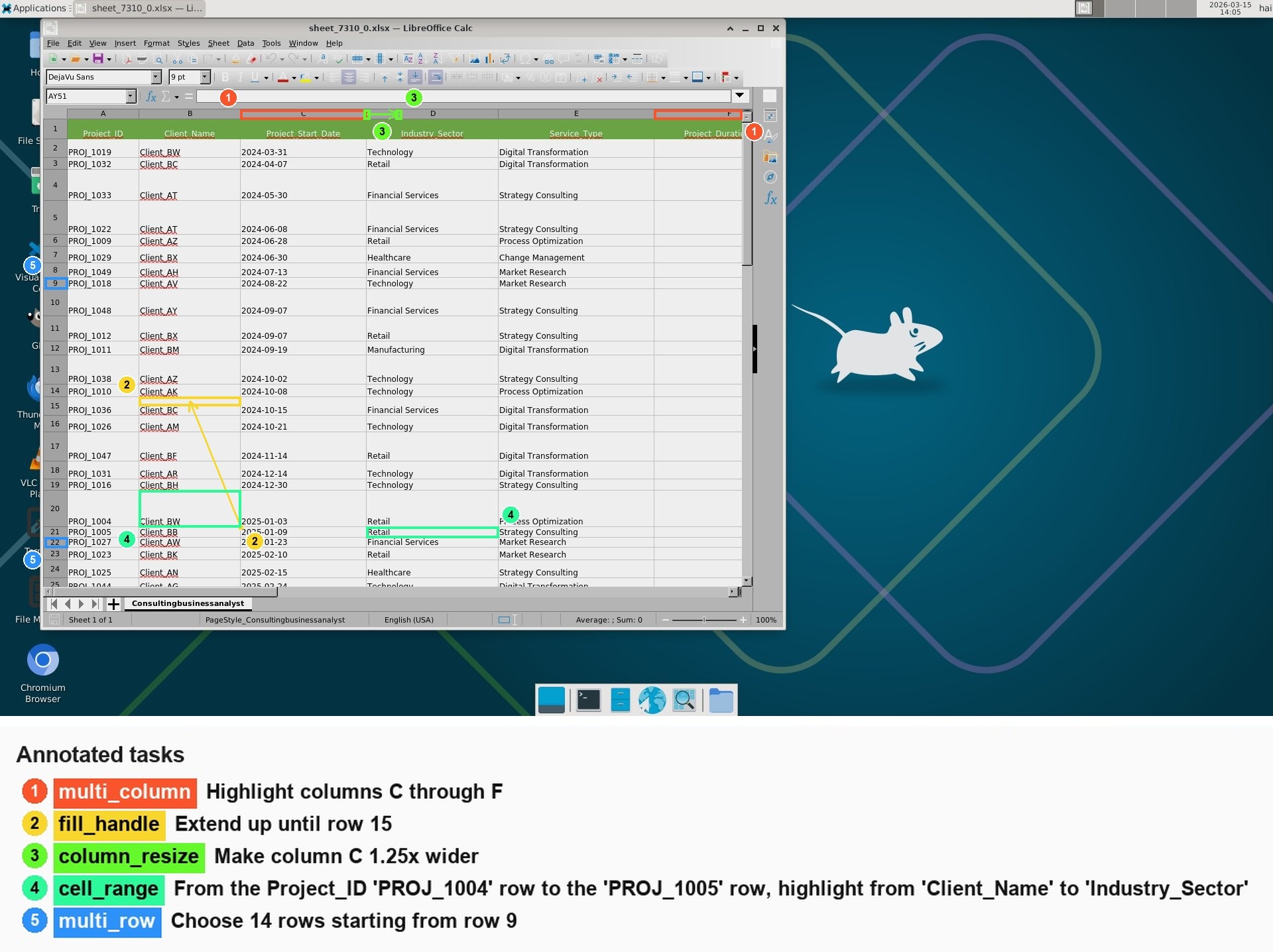}
    \caption{\textit{Cell selection}}
    \label{fig:teaser-sheet}
  \end{subfigure}

  \vspace{0.6em}

  \begin{subfigure}[b]{0.495\textwidth}
    \centering
    \includegraphics[width=\linewidth]{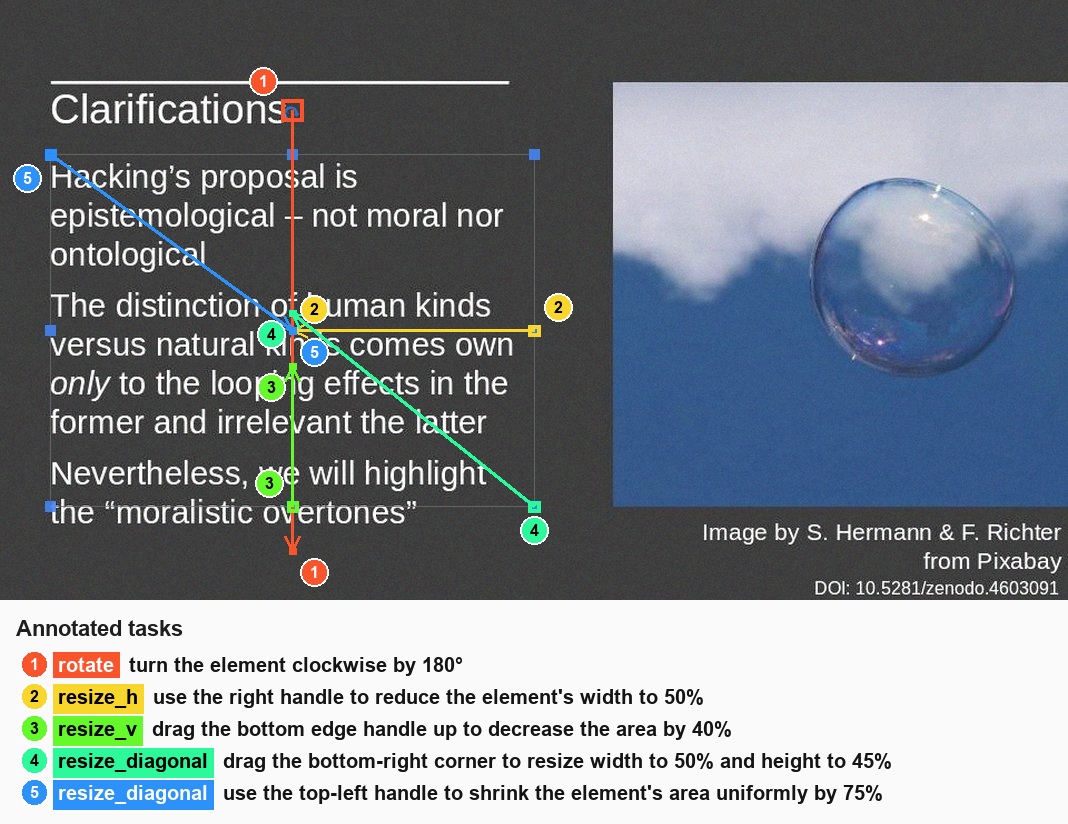}
    \caption{\textit{Element resizing}}
    \label{fig:teaser-slide}
  \end{subfigure}
  \hfill
  \begin{subfigure}[b]{0.495\textwidth}
    \centering
    \includegraphics[width=\linewidth]{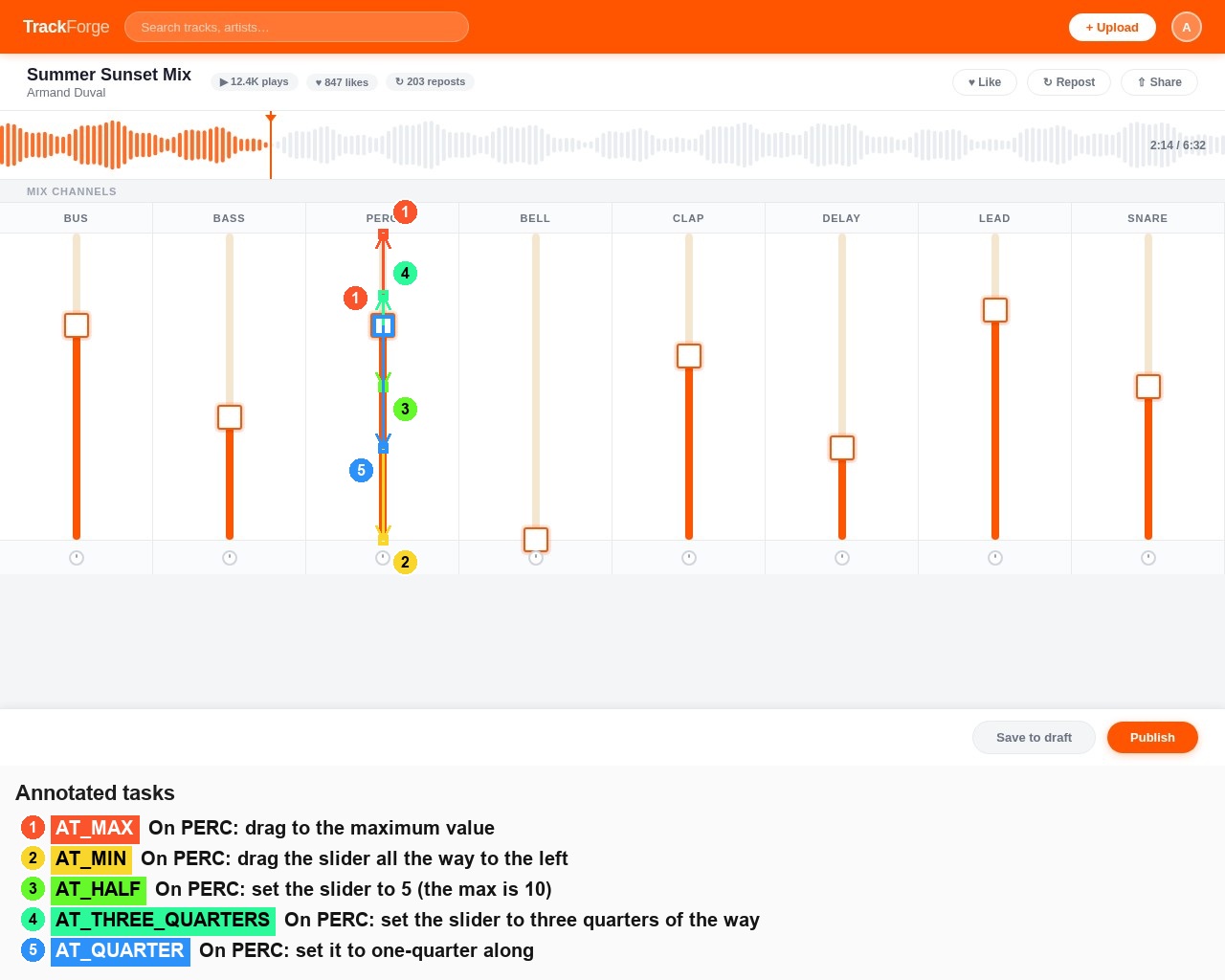}
    \caption{\textit{Slider manipulation}}
    \label{fig:teaser-slider}
  \end{subfigure}

  \caption{The four drag grounding action domains covered by the proposed DragOn benchmark. Each example pairs a screenshot with a natural-language intent; the task is to
  predict a source and target bounding box on the screenshot, with an \texttt{ordered} flag indicating whether drag direction is semantically meaningful.}
  \label{fig:teaser}
\end{figure*}

Recent advances in Large Language Models (LLMs) and Vision-Language Models (VLMs) have substantially improved the reasoning capabilities required for
agentic applications \cite{zhou_mai-ui_2025, andreux_surfer_2025, wang_ui-tars-2_2025}. 
Many real-world software systems and websites expose neither a public API nor a Model Context Protocol (MCP) endpoint, making the graphical user interface the only available control surface.
Thus, visual grounding - the task of mapping a natural language instruction to an actionable screen region - plays a central role in the development of autonomous agents 
capable of performing complex tasks.

Early work on visual grounding has focused mainly on click grounding, which refers to a VLM ability to output a single coordinate (typically to execute a click action in a GUI).
Click grounding has benefited substantially from scale: OS-ATLAS \cite{wu_os-atlas_2024} released over 13M GUI elements across five platforms, UGround \cite{gou_navigating_2024} assembled
10M elements over 1.3M screenshots, and ScreenSpot \cite{cheng_seeclick_2024} established a canonical evaluation on which subsequent models show steady progress. 
In contrast, less focus has been paid to \emph{drag grounding}, the ability to output a pair of coordinates (e.g. to execute swipes or drag-and-drop actions in a GUI).
For drag grounding, available data has remained an order of magnitude smaller \cite{liao_beyond_2025,hu_showui-pi_2025} and largely dominated by text highlighting tasks.
As a result, we show that current VLMs continue to perform poorly on drag-based interactions that are pervasive in GUI workflows.

In this paper, we close this gap by providing a large-scale drag grounding benchmark and training sets. Our contributions are threefold:

\vspace{-5pt}
\paragraph{Rendering-as-supervision.} We formalize a principle for constructing GUI-grounding data by exploiting the renderer's (PDF, XLSX, PPTX, HTML) own geometry as the labeling function. This method yields annotations at a fraction of the cost of VLM- and OCR-based approaches, with fewer construction errors.

\vspace{-5pt}
\paragraph{DragOn Dataset.} We instantiate the principle on four heterogeneous drag grounding domains -- text highlighting, cell selection, element resizing, and slider manipulation -- yielding a unified corpus of 286K training screenshots and 3.5M tasks, plus a 2k-example held-out evaluation suite. The dataset is one to two orders of magnitude larger than prior drag corpora.

\vspace{-5pt}
\paragraph{Benchmark and baselines.}
We evaluate proprietary (GPT, Claude, Gemini) and open-source (Qwen, Kimi, Holo) models, as well as a VLM finetuned on our training data.
Frontier models all score below 30\%, showing that drag grounding remains a challenging problem for current VLMs. We also fine-tune a Qwen3.5VL on our training set, outperforming all frontier models on this task and
showing that additional data can help improve VLM performance on drag actions.

\section{Related Work}

A growing body of work investigates general-purpose agents that perceive, reason about, and act within GUIs, treating visual interaction as a universal interface for autonomous computer-use.

\vspace{-5pt}
\paragraph{Computer-Use Agents.}
MAI-UI \cite{zhou_mai-ui_2025} introduces a family of foundation GUI agents trained with a self-evolving data pipeline and an online reinforcement learning framework, reaching state-of-the-art results on grounding and mobile navigation benchmarks.
Surfer 2 \cite{andreux_surfer_2025} proposes a unified, vision-only architecture that combines hierarchical context management, decoupled planning and execution, and self-verification to operate across web, desktop, and mobile environments without task-specific fine-tuning.
UI-TARS-2 \cite{wang_ui-tars-2_2025} further advances native GUI agents through multi-turn reinforcement learning and a hybrid environment integrating file systems and terminals, demonstrating strong performance across GUI, game, and software engineering tasks.

\vspace{-5pt}
\paragraph{Agent Benchmarks.}
To evaluate performance of these agents, several benchmarks have been proposed.
OSWorld \cite{xie_osworld_2024} provides a real, executable computer environment spanning Ubuntu and Windows, with 369 open-domain tasks evaluated by execution-based scripts; it highlights a large gap between human and model performance, attributed primarily to GUI grounding and operational knowledge.
AndroidWorld \cite{rawles_androidworld_2025} offers a fully functional Android environment with reward signals over 116 programmatic tasks across 20 real-world apps, parameterized to yield diverse natural-language variations of each task.
WebVoyager \cite{he_webvoyager_2024} introduces an end-to-end web agent benchmark built from real-world tasks across 15 popular websites, together with an automatic evaluation protocol leveraging a multimodal model as judge.
While these benchmarks measure overall task success, they provide limited insight into the specific perception and action subtasks on which agents fail.
Drag interactions appear recurrently on end-to-end benchmarks.

\begin{figure*}[t]
  \centering
  \begin{subfigure}[t]{0.495\textwidth}
    \centering
    \includegraphics[width=\linewidth]{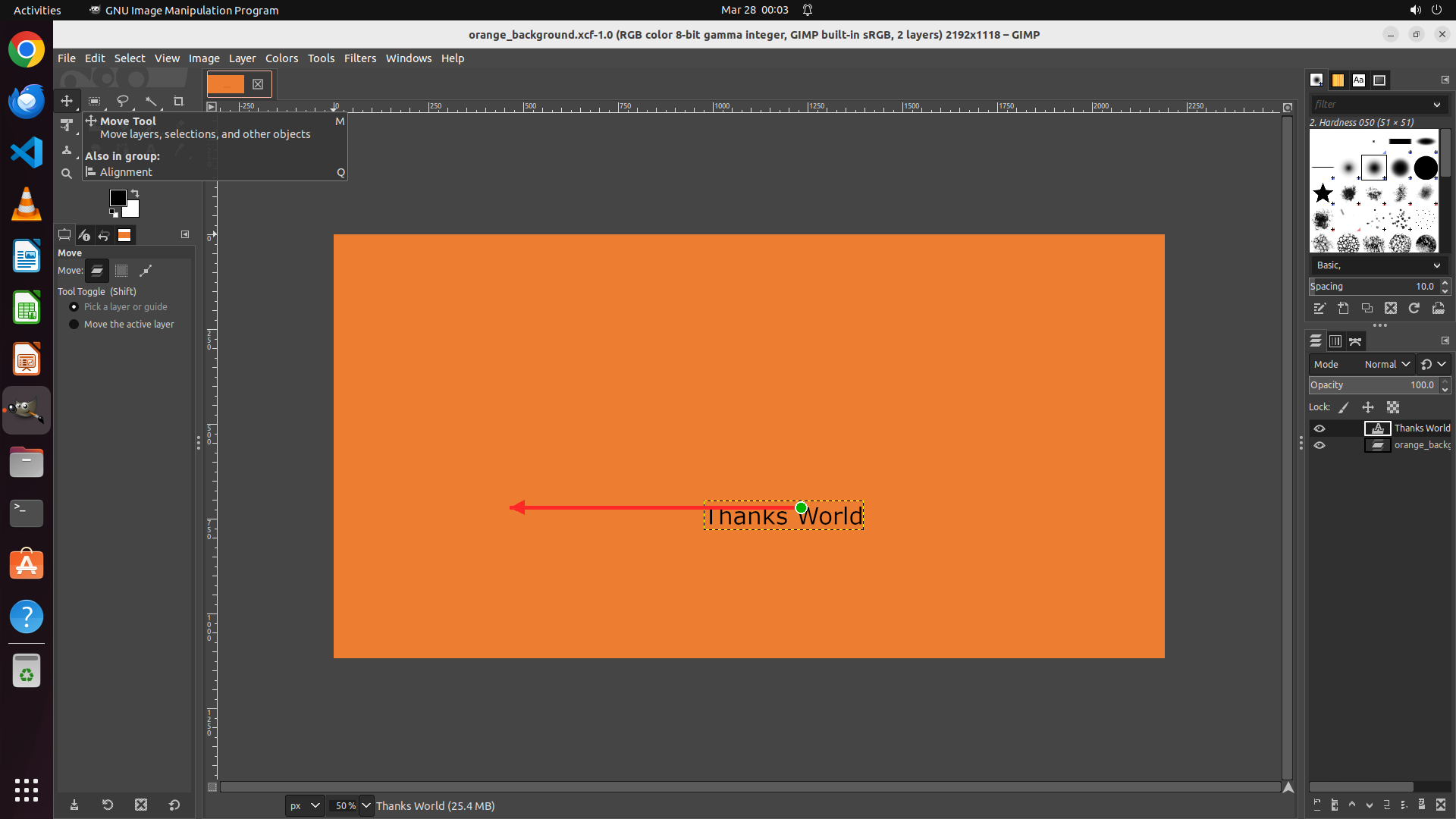}
    \caption{OSWorld (\texttt{gimp\_19}, step~2). Task instruction: \emph{``Can you assist me in shifting the text box to the left? I keep accidentally selecting the image layer beneath it.''} The agent issues \texttt{drag\_and\_drop(x1=0.55, y1=0.62, x2=0.35, y2=0.62)} to drag the `Thanks World' text box leftward on the canvas.}
    \label{fig:agentic-drag-osw}
  \end{subfigure}
  \hfill
  \begin{subfigure}[t]{0.495\textwidth}
    \centering
    \includegraphics[width=\linewidth]{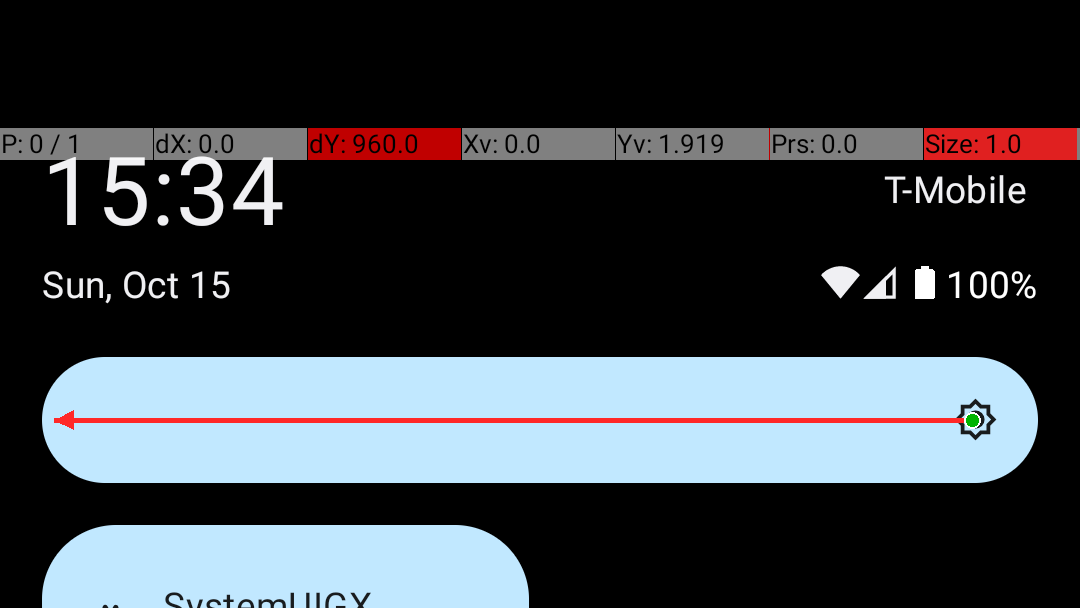}
    \caption{AndroidWorld (\texttt{SystemBrightnessMin-0}, step~2). Task instruction: \emph{``Turn brightness to the min value.''} The agent issues \texttt{mobile\_drag(x1=0.9, y1=0.175, x2=0.05, y2=0.175)} to sweep the brightness slider down to the minimum. (Screenshot cropped to the top region containing the action.)}
    \label{fig:agentic-drag-aw}
  \end{subfigure}

  \caption{Representative drag actions from end-to-end agent benchmarks. The red arrow shows the executed drag overlaid on the observation the agent saw immediately before acting; drag endpoints are in normalized $(x,y)$ screen coordinates.}
  \label{fig:agentic-drag}
\end{figure*}

Indeed, we find that $50/361$ ($13.9\%$) of OSWorld~\cite{xie_osworld_2024} tasks and $96/116$ ($82.8\%$) of AndroidWorld~\cite{rawles_androidworld_2025} tasks require at least
one drag action in a successful trajectory.
On desktop, these drags cluster in spreadsheet range selection, text highlighting, as well as element resizing.
On mobile, drags are near ubiquitous because simply launching an app from the home screen typically requires a vertical swipe on the app drawer. They are also used as a general-purpose swipe primitive across a variety of apps: freehand drawing in browser canvases (e.g.\ \texttt{BrowserDraw}, \texttt{BrowserMultiply}), map interaction in OsmAnd (placing markers, editing tracks), and direct manipulation of UI controls such as the system brightness slider.
Appendix~\ref{sec:appendix-agent-drag-trials} lists every affected trial. Figure~\ref{fig:agentic-drag} shows representative drags from each benchmark.

\vspace{-5pt}
\paragraph{Grounding Datasets.}
A complementary line of work isolates the grounding subtask, namely localizing the screen element referred to by a natural language instruction, which is a prerequisite for any successful action.
SeeClick \cite{cheng_seeclick_2024} identifies grounding as a key bottleneck for vision-only GUI agents and introduces ScreenSpot, a grounding benchmark covering mobile, desktop, and web interfaces, together with a dedicated grounding pre-training procedure.
OS-ATLAS \cite{wu_os-atlas_2024} releases the largest open-source cross-platform grounding corpus to date, with over 13 million GUI elements collected across Windows, Linux, macOS, Android, and the web, and trains a foundation action model that substantially narrows the gap with proprietary VLMs on grounding and out-of-distribution tasks.
UGround \cite{gou_navigating_2024} trains a universal visual grounding model on 10M GUI elements over 1.3M screenshots, showing that a pure-vision agent matches or exceeds systems relying on accessibility-tree input.
ScreenSpot-Pro \cite{li_screenspot-pro_2025} extends grounding evaluation to high-resolution professional applications with small targets and complex layouts, where existing models perform poorly.
More recently, Aria-UI \cite{yang_aria-ui_2025} and GUI-Actor \cite{wu_gui-actor_2025} have further advanced click grounding through pure-vision instruction tuning and a coordinate-free attention-based action head, respectively.
These datasets and methods, however, focus almost exclusively on click-grounding and do not evaluate the continuous, drag-based interactions that we target in this work.

\vspace{-5pt}
\paragraph{Programmatic and Synthetic Supervision.}
A complementary line of work constructs training data through structured pipelines rather than human annotation. Snorkel \cite{ratner_snorkel_2017} formalizes weak supervision through user-written labeling functions whose noise is resolved by a generative model. Domain randomization \cite{tobin_domain_2017} randomizes the renderer's parameters to diversify training inputs for sim-to-real transfer; its contribution lies in the input distribution rather than in the labeling channel. The closest precedents for our use of the renderer as an exact label source come from document understanding: DocBank \cite{li_docbank_2020} extracts token-level bounding boxes from compiled LaTeX source, and Donut \cite{kim_ocr-free_2022} renders synthetic documents to produce paired image-text data at scale. We share their use of the renderer as oracle, but where these pipelines fix a single task per renderer (read all the text, extract all the layout boxes), our label map $\pi$ is \emph{query-conditional} and open-ended over substructures of $S$: the same $(R, S)$ pair supports an arbitrary family of $\pi$ at character, word, span, or layout granularity, parameterized by the natural-language query.

\vspace{-5pt}
\paragraph{Drag Grounding.}
Two recent works target drag-based interactions specifically.
\cite{liao_beyond_2025} introduce a training dataset and benchmark focusing exclusively on text selection. Our work subsumes text highlighting as one of four action domains and uses PDF coordinate extraction rather than OCR, yielding pixel-exact labels at arbitrary granularity - from single characters to multi-paragraph spans.
\cite{hu_showui-pi_2025} propose a flow-based generative model for continuous drag trajectories, trained on 20K trajectories and evaluated on a 505-example benchmark split evenly across five application domains: PowerPoint, Adobe Premiere Pro, an OS file manager, handwriting, and captcha rotation. Their formulation targets closed-loop trajectory prediction whereas we focus on two-point bounding-box grounding from a static screenshot, a more tractable and faster-to-evaluate problem. 

Relative to both prior works, our dataset is one to two orders of magnitude larger (see Table~\ref{tab:related-scale}) and spans two unexplored domains: \textit{cell selection}, and \textit{slider manipulation}.

\begin{table*}[t]
\centering
\caption{Scale and scope of GUI-grounding corpora. Our dataset brings click-grounding scale to drag, across four heterogeneous action domains.}
\label{tab:related-scale}
\setlength{\tabcolsep}{10pt}
\begin{tabular}{@{}lcccc@{}}
\toprule
\textbf{Dataset} & \textbf{Modality} & \textbf{Train} & \textbf{Eval} & \textbf{Domains} \\
\midrule
OS-ATLAS~\cite{wu_os-atlas_2024}         & click           & 13M       & ---            & cross-platform \\
GUI-Drag~\cite{liao_beyond_2025}         & text drag       & 161K      & 5{,}333        & 1 (text)       \\
ShowUI-$\pi$~\cite{hu_showui-pi_2025}    & drag trajectory & 20K       & ---            & 5 apps         \\
\midrule
\textit{DragOn} (ours)          & drag grounding  & 3{,}5M & 2{,}000 & 4 domains \\
\bottomrule
\end{tabular}
\end{table*}

\section{Rendering-as-Supervision}

The four pipelines that produce \textit{DragOn} share a common construction principle. Rather than annotating screenshots, we read ground-truth source and target bounding boxes directly from the renderer geometry. We refer to this principle as \textit{rendering-as-supervision} and use it as a unifying frame for the dataset.

\subsection{Principle}

We use $R$ to denote a deterministic renderer that maps a structured source $S$ to an image $I = R(S)$, and $\pi$ to denote a \emph{label map} that returns a pixel-space answer  from a source and a query: $B = \pi(S, Q)$.

\subsection{Analytic and Probe-Based Label Maps}

The four DragOn domains (see section~\ref{sec:domains}) use two flavors of $\pi$. 
When available, an \emph{analytic} label map computes bounding boxes directly from the intent -- for example, by composing PDF text-span coordinates from source, or by mapping PPTX EMU units to the pixel space of the rendered slide. 
Otherwise, we use \emph{probe-based} label map that runs the renderer twice: a probe pass on a perturbed source (e.g.\ a cell flooded with a distinctive color key) is used to localize the target visually, after which the original source is re-rendered for the final screenshot.
Probe-based label map is needed when no closed-form mapping is available but the rendered geometry can still be recovered to within rendering tolerance through visual detection on a controlled perturbation, which is the case in our cell selection domain (see section~\ref{sec:cell}).

\section{Benchmark}

We create \textit{DragOn}, a benchmark covering four types of complex drag actions: \textit{text highlighting}, \textit{cell selection}, \textit{element resizing} and \textit{slider manipulation}. In total it comprises 286K training screenshots and 3.5M training tasks, together with a 2{,}000-example held-out suite split evenly between a public validation set and a held-out test set (see \Cref{tab:dataset-stats}). The four domains differ only in their instantiation of the $(S, R, \pi)$ triple: PDF span coordinates, color-key pixel detection on a LibreOffice viewport, PPTX EMU units, and URL-parameterized HTML slider geometry, respectively. We summarize them below.

\begin{table*}[t]
\centering
\caption{Dataset statistics. Train counts reflect the unified benchmark after per-image task caps (10/26/12/10 for text/sheet/slide/slider) are applied to the upstream source corpora. Test and eval are disjoint held-out splits with one canonical intent per screenshot.}
\label{tab:dataset-stats}
\small
\begin{tabular}{@{}lrrrrl@{}}
\toprule
\textbf{Domain} & \textbf{Train imgs} & \textbf{Train tasks} & \textbf{Test} & \textbf{Eval} & \textbf{Dominant resolution(s)} \\
\midrule
Text Highlighting & 100{,}000 & 1{,}000{,}000 & 250  & 250  & $1275\!\times\!1650$ (portrait) \\
Cell Selection    &  38{,}332 &   996{,}632   & 250  & 250  & $1920\!\times\!1080$ \\
Element Resizing  &  82{,}680 &   992{,}160   & 250  & 250  & $1280\!\times\!720$, $960\!\times\!720$, $960\!\times\!540$ \\
Slider            &  65{,}000 &   571{,}350   & 250  & 250  & $1280\!\times\!800$ \\
\midrule
\textbf{Total}    & \textbf{286{,}012} & \textbf{3{,}560{,}142} & \textbf{1{,}000} & \textbf{1{,}000} & --- \\
\bottomrule
\end{tabular}
\end{table*}

\subsection{Domain Instantiations}
\label{sec:domains}

Each of the four domains is generated through a dedicated pipeline that produces (i) a screenshot, (ii) a natural-language instruction, and (iii) the source and target bounding boxes defining the drag trajectory. Across all domains, bounding boxes are pixel-aligned with the rendered image either through analytic coordinate extraction or through probe-based visual detection, as described above. Light image augmentations (e.g.\ JPEG compression, blur, brightness and contrast jitter, small rotations) are applied to improve robustness.

Each instruction is sampled from a template that refers to the target by its: textual content, position (absolute or relative to another element), visual style, named entity, structural or semantic role (e.g. column header or a resize handle) or a quantitative specification (e.g. a rotation angle or a slider value). Most real instructions combine several of these.
To ensure every instruction is answerable from the screenshot alone, we check referential uniqueness against the rendering before emission: ambiguous candidates are either scoped (\textit{``in paragraph 3, select `Federal'\,''}), disambiguated with an ordinal (\textit{``the second `Federal' on the page''}), or dropped. Appendix~\ref{sec:appendix-intents} shows example screenshots paired with the full set of intents sampled for each.

\vspace{-5pt}
\paragraph{Text Highlighting.}
\emph{$S$: styled DOCX from a Wikipedia article. $R$: headless LibreOffice DOCX-to-PDF. $\pi$: analytic, via PyMuPDF span coordinates and the page-to-image affine map.}
We rely on Kaggle's structured Wikipedia dump as a static source of English and French articles. Each article section is parsed into plain text and rewritten as a styled DOCX document with varied fonts, colors, and inline highlights, so as to mimic the visual diversity of real-world documents. The DOCX files are converted to PDF using a headless LibreOffice instance, and the PDF is treated as the rendering source of truth for all downstream processing. Text selections of varying granularity (character, word, sentence, paragraph) are sampled with simple heuristics and combined with templated natural-language intents. Bounding boxes for the selected spans are obtained directly from PDF coordinate extraction, which yields exact pixel-aligned annotations on the rasterized image. Contrary to \cite{liao_beyond_2025}, which relies on OCR to recover word-level boxes, extracting span coordinates directly from the PDF lets us vary the granularity at will, supporting tasks that require highlighting at the character level as well as at the paragraph level.

\vspace{-5pt}
\paragraph{Cell Selection.} \label{sec:cell}
\emph{$S$: XLSX from a Pandas DataFrame. $R$: LibreOffice Calc in a Debian Docker image. $\pi$: probe-based, via color-key detection on a perturbed viewport.}
Spreadsheet examples are fully synthetic. Realistic-looking tables, including calendars and gradebooks, are generated on-the-fly from Pandas DataFrames, exported to XLSX, and opened in a real LibreOffice Calc instance running inside a Debian Docker image. To recover pixel-accurate cell positions in the rendered viewport, we temporarily fill each cell with a distinctive color key, capture a screenshot, and locate cells by color matching, before reverting to the original styling for the final screenshot. Instructions are then templated from the detected cell grid and cover a range of interactions, including rectangular cell-range selections, multi-row and multi-column selections, column- and row-border resizes, and fill-handle drags (e.g.\ ``extend formula down").

\vspace{-5pt}
\paragraph{Element Resizing.}
\emph{$S$: a slide from the Zenodo 10K PowerPoint corpus. $R$: headless LibreOffice slide-to-JPEG. $\pi$: analytic, via the EMU-to-pixel affine and per-handle offsets.}
Slide examples are derived from the Zenodo 10K corpus of real-world PowerPoint presentations~\cite{datacite_data_2025}. Each deck is split into individual slides, which are rendered to JPEG via headless LibreOffice. Shape geometry is extracted directly from the underlying .pptx XML using \texttt{python-pptx} in native EMU units \footnote{English Metric Unit: A measurement in computer typography. There are 635 EMUs per twip, 6,350 EMUs per half-point, 12,700 EMUs per point, and 914,400 EMUs per inch. Used to translate on-screen layouts to printed layouts for specified printer hardware.}, and then mapped to the pixel space of the rendered slide. For each shape, we sample resize, rotate, and crop actions, each parameterized by one of the eight standard manipulation handles (four corners, four mid-edges, plus a rotation handle) and a target drag destination. Bounding boxes are computed geometrically from the shape's EMU bounding box and the corresponding handle offsets, so no visual detection is required. Annotated images additionally overlay the shape's selection border and handle circles, as illustrated in our training data.

\vspace{-5pt}
\paragraph{Slider Manipulation.}
\emph{$S$: a URL-parameterized HTML page. $R$: headless Playwright-Chromium. $\pi$: analytic, from value, track geometry, and per-variant thumb dimensions.}
Slider examples are procedurally generated, with no external source data. We define six interactive HTML contexts, including a card dashboard, a media player, a settings panel, a photo editor, an audio mixer, and an e-commerce page, each instantiated in three visual variants. The full parameterization (color, theme, track size, thumb shape and size, value range, step, endpoint labels, variant, etc.) yields a combinatorial space of approximately 850K unique page configurations.

For each URL, we compute which cues are actually visible (e.g.\ endpoint labels, tooltip, volume-icon convention with a 0–100 range) and sample instructions from a catalog of 17 templates restricted to those present in the rendering. This ensures every instruction can be answered from the screenshot alone. Pages are rendered with a headless Playwright-Chromium browser, captured at the slider initial value, and saved as JPEG. The source bounding box is derived from the thumb pixel position, computed from the initial and target values together with the track geometry and per-variant thumb dimensions; the target bounding box is a fixed 10$\times$10 region around the drop point. Train, validation, and test splits are assigned deterministically by MD5 bucketing on the URL, ensuring that identical configurations consistently fall in the same split across regenerations.

\subsection{Difficulty and Tolerance}

Both resize and rotation are challenging tasks because the intent is highly precise (e.g.\ ``increase the width of X by 30\% of its current size'' for resize, or ``rotate X by 45 degrees clockwise'' for rotation) and we set the target bounding-box dimensions to 5\% of the element’s current height and width (i.e. tolerance of $\pm2.5\%$).
To better discriminate between models, we additionally report performance under a relaxed tolerance on the target bounding box: in the following, \textit{acc@10\%} and \textit{acc@15\%} denote success when the predicted point falls within a region scaled by 10\% and 15\% of the slide dimensions around the target, respectively.

\subsection{Degrees of Freedom}

For both resize and rotation, the set of acceptable target locations is in fact a continuum rather than a single point.
When resizing along a single dimension using an edge handle, any point on the line orthogonal to that dimension passing through the target is equally valid.
Similarly, for the 2D rotations considered in this work, any point on the half-line originating at the element center with the prescribed angle yields the same rotation.
We illustrate the two conventions side-by-side in Figure~\ref{fig:bbox-alternatives} in the appendix.

For simplicity, we ignore these degrees of freedom and define the target bounding box canonically: for resize, on the axis aligned with the starting handle; for rotation, on the circle centered at the element center with radius equal to the distance from the starting handle to that center, on the same axis as the starting handle.
Crucially, the training set is generated using the same canonical convention, preventing any spurious mismatch between training and evaluation targets.

\subsection{Ordering}

Some tasks require the click-and-drag to be executed in a specific direction, while others are direction-agnostic. In particular, cell selection and text highlighting accept either ordering of the two endpoints: dragging from the source to the target and dragging from the target to the source produce the same selection. In contrast, element resizing, rotation, and slider manipulation are inherently ordered, since the action is defined relative to the initial handle position. We expose this property explicitly via the \texttt{"ordered"} boolean field in the per-example JSON, which is used by the evaluation script.
See figures \ref{fig:bbox-alternatives-resize} and \ref{fig:bbox-alternatives-rotate} in appendix for an illustration.

\subsection{Test Sets}

We provide two held-out evaluation sets. The \emph{validation set} is publicly released and intended for development and self-reported results. The \emph{test set} is kept private and serves as the reference benchmark for cross-model comparison; participants may submit predictions and obtain their score, with optional inclusion on a public leaderboard. Both sets contain 1{,}000 examples, evenly distributed across the four task types (250 examples each), with a single canonical intent per screenshot.

\section{Experiments}

We evaluate a diverse set of models on our benchmark, spanning closed-source generalist agents accessed through public APIs and open-weight vision-language models served locally. We first describe the training procedure for our own model, then detail the evaluation protocol used for each baseline.
All models are queried with structured (JSON) outputs encoding the four drag coordinates.

\subsection{Training}

We fine-tune a pretrained Qwen3.5-VL-35B-A3B~\cite{qwen_qwen35_2026} model on our training dataset. We use Adam optimizer~\cite{kingma_adam_2017} with learning rate \texttt{1e-6}, sequence length $20{,}480$, global batch size of $64$ on $32$ Nvidia H100 GPUs. We train the model for $1000$ steps, corresponding to approximately 2 epochs. 

\subsection{Evaluation}

We evaluate generalist and specialized models on the benchmark. For each baseline, we use the recommended settings provided in the documentation (e.g. decoding configuration, native coordinate convention), as described below:

\vspace{-5pt}
\paragraph{Claude Sonnet 4.5.}
We query Claude Sonnet 4.5~\cite{anthropic_claude_2025} through the Anthropic API with extended thinking disabled, and obtain the four drag coordinates as a forced tool call over a JSON schema, with coordinates expressed in image-pixel units.

\vspace{-5pt}
\paragraph{Claude Opus 4.7.}
Claude Opus 4.7~\cite{anthropic_claude_2025} is queried in the same setting as Sonnet 4.5: Anthropic API, extended thinking disabled, structured output through a forced tool call, image-pixel coordinates.

\vspace{-5pt}
\paragraph{GPT-5.4.}
GPT-5.4~\cite{singh_openai_2025} is queried through the OpenAI API in JSON-object mode with default decoding, with coordinates expressed in image-pixel units.

\vspace{-5pt}
\paragraph{Kimi-K2.5.}
Kimi-K2.5~\cite{team_kimi_2026} is queried through the Together AI API in JSON-object mode at temperature $0.6$ and top-$p$ $0.95$ with reasoning disabled, and outputs coordinates as fractions of the image dimensions in $[0, 1]$.

\vspace{-5pt}
\paragraph{Qwen3.5-VL and Holo3-35B-A3B.}
The Qwen3.5-VL-35B-A3B~\cite{qwen_qwen35_2026} and Holo3-35B-A3B~\cite{hcompany_holo3_2026} are served locally with vLLM in JSON-object mode with chain-of-thought mode enabled, and emit coordinates in the Qwen-native normalized $[0, 1000]$ frame. Qwen3.5-VL-122B-A10B and Qwen3.5-VL-397B-A17B are served in FP8 precision.

\subsection{Results}

\begin{table*}[!t]
  \centering
  \caption{Success rate on the test set. The last two columns report success on element resizing under 10\% and 15\% spatial tolerance. Best result per column in \textbf{bold}.}
  \label{tab:results}
  \footnotesize
  \setlength{\tabcolsep}{3pt}
  \begin{tabular}{@{}lccccccc@{}}
    \toprule
    & \multicolumn{5}{c}{\textbf{Success rate (\%)}} & \multicolumn{2}{c}{\textbf{Tolerant ER (\%)}} \\
    \cmidrule(lr){2-6} \cmidrule(l){7-8}
    \textbf{Model} & \textbf{Total} & \textbf{Text Highlighting} & \textbf{Cell Selection} & \textbf{Slider Manipulation} & \textbf{Element Resizing} & \textbf{acc@10\%} & \textbf{acc@15\%} \\
    \midrule
    Claude Sonnet 4.5             & 10.7 &  0.0 &          0.0  & 26.0 & 16.8 & 19.6 & 20.4 \\
    Claude Opus 4.7               & 27.7 &  7.6 & \textbf{37.2} & 57.2 &  8.8 & 13.6 & 16.0 \\
    GPT-5.4                       & 25.7 &  9.2 &         28.0  & 45.6 & 20.0 & 23.2 & 26.4 \\
    Kimi-K2.5                     & 11.6 & 12.4 &         14.0  & 16.0 &  4.0 &  5.2 &  7.6 \\
    Holo3 35B-A3B                 & 23.2 & 20.8 &         34.4  & 14.4 & 23.2 & 30.0 & 36.4 \\
    Qwen3.5 35B-A3B               &  2.3 &  1.6 &          0.4  &  4.8 &  2.4 &  2.8 &  4.0 \\
    Qwen3.5 122B-A10B             & 10.8 &  6.4 &          7.2  & 19.6 & 10.0 & 14.8 & 18.8 \\
    Qwen3.5 397B-A17B             & 21.5 & 16.4 &         16.4  & 39.6 & 13.6 & 19.6 & 24.8 \\
    \midrule
    \textbf{\textit{Ours} 35B-A3B} & \textbf{35.3} & \textbf{33.2} & 13.2 & \textbf{62.8} & \textbf{32.0} & \textbf{46.8} & \textbf{57.2} \\
    \bottomrule
  \end{tabular}
\end{table*}

Results on the test set are reported in Table~\ref{tab:results}, our model being denoted as \textit{ours-35b-a3b}.
The two strongest proprietary models, \textit{claude-opus-4-7} ($27.7\%$) and \textit{gpt-5.4} ($25.7\%$), perform reasonably well on slider manipulation ($57.2\%$ and $45.6\%$) and cell selection ($37.2\%$ and $28.0\%$), but their accuracy drops drastically on text highlighting (both below $10\%$) and element resizing (both below $20\%$). \textit{claude-sonnet-4-5} is the weakest of the proprietary models on our benchmark, with an overall score of $10.7\%$ and near-zero accuracy on the two text-based categories. Overall, no frontier model clears $30\%$, confirming that drag grounding is far from saturated even for the strongest closed systems.

Second, open-weight models trail proprietary frontier at matched or larger parameter counts. \textit{qwen3-5-35b-a3b} collapses at $2.3\%$, \textit{Kimi-K2.5} and \textit{qwen3-5-122b-a10b-fp8} sit around $11\%$, and only the largest variant \textit{qwen3-5-397b-a17b-fp8} ($21.5\%$) approaches the proprietary band, still without reaching it. Computer-use specialization has a large effect at fixed architecture: compared to \textit{qwen3-5-35b-a3b}, \textit{holo3-35b-a3b} (which share the same 35B/A3B Mixture-of-Experts backbone) show an absolute performance boost of $20.9$ points, close to proprietary frontier models.

Furthermore, we show that targeted training on our corpus pushes performance well past the frontier. Fine-tuning Qwen 35B/A3B base on our drag grounding data (\textit{Ours-35b-a3b}) lifts overall success from $2.3\%$ to $35.3\%$, a 33-point absolute improvement. Our model surpasses every evaluated frontier model by at least $7.6$ points and takes the lead on three of the four action domains (text highlighting, slider manipulation, and element resizing), while remaining behind \textit{claude-opus-4-7} on cell selection. These gains come entirely from data: the model class, size, and training pipeline otherwise match the open-source Qwen3-5 baseline.

Finally, a few limits of the current results are worth flagging. Element resizing remains the hardest domain in absolute terms, but the acc@10\% and acc@15\% columns suggest it is tolerance-limited rather than localization-limited: our model improves from $32.0\%$ at the strict threshold to $46.8\%$ at $10\%$ and $57.2\%$ at $15\%$ of slide dimensions, with all frontier models moving in the same direction. This indicates that most remaining errors on resize are small-magnitude offsets around the correct handle rather than qualitative mis-groundings. 

\subsection{Code}
\label{sec:code}

Code used to evaluate the models can be found in the following \href{https://github.com/hcompai/DragOn}{Github repository}.
Datasets can be found in the following \href{https://huggingface.co/datasets/HCompany/DragOn}{HuggingFace repository}, which will also include a leaderboard.

\section{Conclusion}

We presented a benchmark and training dataset for drag-based computer-use interactions, an aspect of GUI grounding that has so far received much less attention than click-grounding. The dataset spans four representative domains -- \textit{text highlighting}, \textit{cell selection}, \textit{element resizing}, and \textit{slider manipulation} -- and is built with task-specific pipelines that produce pixel-aligned source and target bounding boxes from structured sources, avoiding the noise of OCR- or detection-based annotation. In total, the release comprises 286K training screenshots and 3.5M training tasks, together with a 2{,}000-example held-out evaluation suite split into a public validation set and a private test set.

A comprehensive set of VLMs were evaluated on the benchmark: proprietary models (GPT, Claude), open-source VLMs (Qwen, Kimi, Holo), and an open-source VLM finetuned on our training data. Our results show that drag grounding remains challenging for current VLMs and that an open-source model fine-tuned on our dataset outperforms the strongest proprietary baselines, indicating that targeted training data could improve model performance on downstream computer-use tasks.

We hope that the benchmark, the dataset, and the public leaderboard will support further progress on drag grounding, a capability required for reliable computer-use agents.

% \section*{Acknowledgements}

\section*{Impact Statement}

This paper presents work whose goal is to advance the field of Machine Learning. There are many potential societal consequences of our work, none of which we feel must be specifically highlighted here.

\section*{LLM and Agent Usage}

We used large language models to proofread the manuscript. Coding agents assisted at various stages of dataset generation, particularly in creating the HTML for the slider selection domain. Figure~\ref{fig:teaser} was also generated using a coding agent, based on actual images and annotations from our dataset.

% \clearpage

\bibliography{main}

@misc{zhou_mai-ui_2025,
	title = {{MAI}-{UI} Technical Report: Real-World Centric Foundation {GUI} Agents},
	url = {http://arxiv.org/abs/2512.22047},
	doi = {10.48550/arXiv.2512.22047},
	shorttitle = {{MAI}-{UI} Technical Report},
	number = {{arXiv}:2512.22047},
	publisher = {{arXiv}},
	author = {Zhou, Hanzhang and Zhang, Xu and Tong, Panrong and Zhang, Jianan and Chen, Liangyu and Kong, Quyu and Cai, Chenglin and Liu, Chen and Wang, Yue and Zhou, Jingren and Hoi, Steven},
	urldate = {2026-01-01},
	date = {2025-12-26},
	eprinttype = {arxiv},
	eprint = {2512.22047 [cs]},
	note = {version: 1},
}

@misc{liao_beyond_2025,
	title = {Beyond Clicking: A Step Towards Generalist {GUI} Grounding via Text Dragging},
	url = {http://arxiv.org/abs/2601.06031},
	doi = {10.48550/arXiv.2601.06031},
	shorttitle = {Beyond Clicking},
	number = {{arXiv}:2601.06031},
	publisher = {{arXiv}},
	author = {Liao, Zeyi and Lu, Yadong and Gou, Boyu and Sun, Huan and Awadallah, Ahmed},
	urldate = {2026-04-21},
	date = {2025-11-07},
	eprinttype = {arxiv},
	eprint = {2601.06031 [cs]},
}

@misc{hu_showui-pi_2025,
	title = {{ShowUI}-pi: Flow-based Generative Models as {GUI} Dexterous Hands},
	url = {http://arxiv.org/abs/2512.24965},
	doi = {10.48550/arXiv.2512.24965},
	shorttitle = {{ShowUI}-\$π\$},
	number = {{arXiv}:2512.24965},
	publisher = {{arXiv}},
	author = {Hu, Siyuan and Lin, Kevin Qinghong and Shou, Mike Zheng},
	urldate = {2026-04-21},
	date = {2025-12-31},
	eprinttype = {arxiv},
	eprint = {2512.24965 [cs]},
}

@misc{xie_osworld_2024,
	title = {{OSWorld}: Benchmarking Multimodal Agents for Open-Ended Tasks in Real Computer Environments},
	url = {http://arxiv.org/abs/2404.07972},
	doi = {10.48550/arXiv.2404.07972},
	shorttitle = {{OSWorld}},
	number = {{arXiv}:2404.07972},
	publisher = {{arXiv}},
	author = {Xie, Tianbao and Zhang, Danyang and Chen, Jixuan and Li, Xiaochuan and Zhao, Siheng and Cao, Ruisheng and Hua, Toh Jing and Cheng, Zhoujun and Shin, Dongchan and Lei, Fangyu and Liu, Yitao and Xu, Yiheng and Zhou, Shuyan and Savarese, Silvio and Xiong, Caiming and Zhong, Victor and Yu, Tao},
	urldate = {2026-04-21},
	date = {2024-05-30},
	eprinttype = {arxiv},
	eprint = {2404.07972 [cs]},
}

@misc{andreux_surfer_2025,
	title = {Surfer 2: The Next Generation of Cross-Platform Computer Use Agents},
	url = {http://arxiv.org/abs/2510.19949},
	doi = {10.48550/arXiv.2510.19949},
	shorttitle = {Surfer 2},
	number = {{arXiv}:2510.19949},
	publisher = {{arXiv}},
	author = {Andreux, Mathieu and Bakler, Märt and Barbier, Yanael and Benchekroun, Hamza and Biré, Emilien and Bonnet, Antoine and Bordie, Riaz and Bout, Nathan and Brunel, Matthias and Cambray, Aleix and Cedoz, Pierre-Louis and Chassang, Antoine and Cloix, Gautier and Connelly, Ethan and Constantinou, Alexandra and Coster, Ramzi De and Jonquiere, Hubert de la and Delfosse, Aurélien and Delpit, Maxime and Deprez, Alexis and Derupti, Augustin and Diaz, Mathieu and D'Souza, Shannon and Dujardin, Julie and Edmund, Abai and Eickenberg, Michael and Fatalot, Armand and Felissi, Wissem and Herring, Isaac and Koegler, Xavier and Kergaradec, Erwan Le Jumeau de and Lac, Aurélien and Langevin, Maxime and Lauverjat, Corentin and Loison, Antonio and Manevich, Avshalom and Moyal, Axel and Kerbel, Axel Nguyen and Parovic, Marinela and Revelle, Julien and Richard, Guillaume and Richter, Mats and Riochet, Ronan and Santos, María and Savidan, Romain and Sifre, Laurent and Theillard, Maxime and Thibault, Marc and Valentini, Ivan and Wu, Tony and Yie, Laura and Yuan, Kai and Zubovskij, Jevgenij},
	urldate = {2026-04-21},
	date = {2025-10-24},
	eprinttype = {arxiv},
	eprint = {2510.19949 [cs]},
}

@misc{he_webvoyager_2024,
	title = {{WebVoyager}: Building an End-to-End Web Agent with Large Multimodal Models},
	url = {http://arxiv.org/abs/2401.13919},
	doi = {10.48550/arXiv.2401.13919},
	shorttitle = {{WebVoyager}},
	number = {{arXiv}:2401.13919},
	publisher = {{arXiv}},
	author = {He, Hongliang and Yao, Wenlin and Ma, Kaixin and Yu, Wenhao and Dai, Yong and Zhang, Hongming and Lan, Zhenzhong and Yu, Dong},
	urldate = {2026-04-22},
	date = {2024-06-06},
	eprinttype = {arxiv},
	eprint = {2401.13919 [cs]},
}

@misc{rawles_androidworld_2025,
	title = {{AndroidWorld}: A Dynamic Benchmarking Environment for Autonomous Agents},
	url = {http://arxiv.org/abs/2405.14573},
	doi = {10.48550/arXiv.2405.14573},
	shorttitle = {{AndroidWorld}},
	number = {{arXiv}:2405.14573},
	publisher = {{arXiv}},
	author = {Rawles, Christopher and Clinckemaillie, Sarah and Chang, Yifan and Waltz, Jonathan and Lau, Gabrielle and Fair, Marybeth and Li, Alice and Bishop, William and Li, Wei and Campbell-Ajala, Folawiyo and Toyama, Daniel and Berry, Robert and Tyamagundlu, Divya and Lillicrap, Timothy and Riva, Oriana},
	urldate = {2026-04-22},
	date = {2025-04-06},
	eprinttype = {arxiv},
	eprint = {2405.14573 [cs]},
}

@misc{wang_ui-tars-2_2025,
	title = {{UI}-{TARS}-2 Technical Report: Advancing {GUI} Agent with Multi-Turn Reinforcement Learning},
	url = {http://arxiv.org/abs/2509.02544},
	doi = {10.48550/arXiv.2509.02544},
	shorttitle = {{UI}-{TARS}-2 Technical Report},
	number = {{arXiv}:2509.02544},
	publisher = {{arXiv}},
	author = {Wang, Haoming and Zou, Haoyang and Song, Huatong and Feng, Jiazhan and Fang, Junjie and Lu, Junting and Liu, Longxiang and Luo, Qinyu and Liang, Shihao and Huang, Shijue and Zhong, Wanjun and Ye, Yining and Qin, Yujia and Xiong, Yuwen and Song, Yuxin and Wu, Zhiyong and Li, Aoyan and Li, Bo and Dun, Chen and Liu, Chong and Zan, Daoguang and Leng, Fuxing and Wang, Hanbin and Yu, Hao and Chen, Haobin and Guo, Hongyi and Su, Jing and Huang, Jingjia and Shen, Kai and Shi, Kaiyu and Yan, Lin and Zhao, Peiyao and Liu, Pengfei and Ye, Qinghao and Zheng, Renjie and Xin, Shulin and Zhao, Wayne Xin and Heng, Wen and Huang, Wenhao and Wang, Wenqian and Qin, Xiaobo and Lin, Yi and Wu, Youbin and Chen, Zehui and Wang, Zihao and Zhong, Baoquan and Zhang, Xinchun and Li, Xujing and Li, Yuanfan and Zhao, Zhongkai and Jiang, Chengquan and Wu, Faming and Zhou, Haotian and Pang, Jinlin and Han, Li and Liu, Qi and Ma, Qianli and Liu, Siyao and Cai, Songhua and Fu, Wenqi and Liu, Xin and Wang, Yaohui and Zhang, Zhi and Zhou, Bo and Li, Guoliang and Shi, Jiajun and Yang, Jiale and Tang, Jie and Li, Li and Han, Qihua and Lu, Taoran and Lin, Woyu and Tong, Xiaokang and Li, Xinyao and Zhang, Yichi and Miao, Yu and Jiang, Zhengxuan and Li, Zili and Zhao, Ziyuan and Li, Chenxin and Ma, Dehua and Lin, Feng and Zhang, Ge and Yang, Haihua and Guo, Hangyu and Zhu, Hongda and Liu, Jiaheng and Du, Junda and Cai, Kai and Li, Kuanye and Yuan, Lichen and Han, Meilan and Wang, Minchao and Guo, Shuyue and Cheng, Tianhao and Ma, Xiaobo and Xiao, Xiaojun and Huang, Xiaolong and Chen, Xinjie and Du, Yidi and Chen, Yilin and Wang, Yiwen and Li, Zhaojian and Yang, Zhenzhu and Zeng, Zhiyuan and Jin, Chaolin and Li, Chen and Chen, Hao and Chen, Haoli and Chen, Jian and Zhao, Qinghao and Shi, Guang},
	urldate = {2026-04-22},
	date = {2025-09-05},
	eprinttype = {arxiv},
	eprint = {2509.02544 [cs]},
}

@misc{li_screenspot-pro_2025,
	title = {{ScreenSpot}-Pro: {GUI} Grounding for Professional High-Resolution Computer Use},
	url = {http://arxiv.org/abs/2504.07981},
	doi = {10.48550/arXiv.2504.07981},
	shorttitle = {{ScreenSpot}-Pro},
	number = {{arXiv}:2504.07981},
	publisher = {{arXiv}},
	author = {Li, Kaixin and Meng, Ziyang and Lin, Hongzhan and Luo, Ziyang and Tian, Yuchen and Ma, Jing and Huang, Zhiyong and Chua, Tat-Seng},
	urldate = {2026-04-22},
	date = {2025-04-04},
	eprinttype = {arxiv},
	eprint = {2504.07981 [cs]},
}

@misc{cheng_seeclick_2024,
	title = {{SeeClick}: Harnessing {GUI} Grounding for Advanced Visual {GUI} Agents},
	url = {http://arxiv.org/abs/2401.10935},
	doi = {10.48550/arXiv.2401.10935},
	shorttitle = {{SeeClick}},
	number = {{arXiv}:2401.10935},
	publisher = {{arXiv}},
	author = {Cheng, Kanzhi and Sun, Qiushi and Chu, Yougang and Xu, Fangzhi and Li, Yantao and Zhang, Jianbing and Wu, Zhiyong},
	urldate = {2026-04-22},
	date = {2024-02-23},
	eprinttype = {arxiv},
	eprint = {2401.10935 [cs]},
}

@misc{singh_openai_2025,
	title = {{OpenAI} {GPT}-5 System Card},
	url = {http://arxiv.org/abs/2601.03267},
	doi = {10.48550/arXiv.2601.03267},
	number = {{arXiv}:2601.03267},
	publisher = {{arXiv}},
	author = {Singh, Aaditya and Fry, Adam and Perelman, Adam and Tart, Adam and Ganesh, Adi and El-Kishky, Ahmed and {McLaughlin}, Aidan and Low, Aiden and Ostrow, A. J. and Ananthram, Akhila and Nathan, Akshay and Luo, Alan and Helyar, Alec and Madry, Aleksander and Efremov, Aleksandr and Spyra, Aleksandra and Baker-Whitcomb, Alex and Beutel, Alex and Karpenko, Alex and Makelov, Alex and Neitz, Alex and Wei, Alex and Barr, Alexandra and Kirchmeyer, Alexandre and Ivanov, Alexey and Christakis, Alexi and Gillespie, Alistair and Tam, Allison and Bennett, Ally and Wan, Alvin and Huang, Alyssa and Sandjideh, Amy {McDonald} and Yang, Amy and Kumar, Ananya and Saraiva, Andre and Vallone, Andrea and Gheorghe, Andrei and Garcia, Andres Garcia and Braunstein, Andrew and Liu, Andrew and Schmidt, Andrew and Mereskin, Andrey and Mishchenko, Andrey and Applebaum, Andy and Rogerson, Andy and Rajan, Ann and Wei, Annie and Kotha, Anoop and Srivastava, Anubha and Agrawal, Anushree and Vijayvergiya, Arun and Tyra, Ashley and Nair, Ashvin and Nayak, Avi and Eggers, Ben and Ji, Bessie and Hoover, Beth and Chen, Bill and Chen, Blair and Barak, Boaz and Minaiev, Borys and Hao, Botao and Baker, Bowen and Lightcap, Brad and {McKinzie}, Brandon and Wang, Brandon and Quinn, Brendan and Fioca, Brian and Hsu, Brian and Yang, Brian and Yu, Brian and Zhang, Brian and Brenner, Brittany and Zetino, Callie Riggins and Raymond, Cameron and Lugaresi, Camillo and Paz, Carolina and Hudson, Cary and Whitney, Cedric and Li, Chak and Chen, Charles and Cole, Charlotte and Voss, Chelsea and Ding, Chen and Shen, Chen and Huang, Chengdu and Colby, Chris and Hallacy, Chris and Koch, Chris and Lu, Chris and Kaplan, Christina and Kim, Christina and Minott-Henriques, C. J. and Frey, Cliff and Yu, Cody and Czarnecki, Coley and Reid, Colin and Wei, Colin and Decareaux, Cory and Scheau, Cristina and Zhang, Cyril and Forbes, Cyrus and Tang, Da and Goldberg, Dakota and Roberts, Dan and Palmie, Dana and Kappler, Daniel and Levine, Daniel and Wright, Daniel and Leo, Dave and Lin, David and Robinson, David and Grabb, Declan and Chen, Derek and Lim, Derek and Salama, Derek and Bhattacharjee, Dibya and Tsipras, Dimitris and Li, Dinghua and Yu, Dingli and Strouse, D. J. and Williams, Drew and Hunn, Dylan and Bayes, Ed and Arbus, Edwin and Akyurek, Ekin and Le, Elaine Ya and Widmann, Elana and Yani, Eli and Proehl, Elizabeth and Sert, Enis and Cheung, Enoch and Schwartz, Eri and Han, Eric and Jiang, Eric and Mitchell, Eric and Sigler, Eric and Wallace, Eric and Ritter, Erik and Kavanaugh, Erin and Mays, Evan and Nikishin, Evgenii and Li, Fangyuan and Such, Felipe Petroski and Peres, Filipe de Avila Belbute and Raso, Filippo and Bekerman, Florent and Tsimpourlas, Foivos and Chantzis, Fotis and Song, Francis and Zhang, Francis and Raila, Gaby and {McGrath}, Garrett and Briggs, Gary and Yang, Gary and Parascandolo, Giambattista and Chabot, Gildas and Kim, Grace and Zhao, Grace and Valiant, Gregory and Leclerc, Guillaume and Salman, Hadi and Wang, Hanson and Sheng, Hao and Jiang, Haoming and Wang, Haoyu and Jin, Haozhun and Sikchi, Harshit and Schmidt, Heather and Aspegren, Henry and Chen, Honglin and Qiu, Huida and Lightman, Hunter and Covert, Ian and Kivlichan, Ian and Silber, Ian and Sohl, Ian and Hammoud, Ibrahim and Clavera, Ignasi and Lan, Ikai and Akkaya, Ilge and Kostrikov, Ilya and Kofman, Irina and Etinger, Isak and Singal, Ishaan and Hehir, Jackie and Huh, Jacob and Pan, Jacqueline and Wilczynski, Jake and Pachocki, Jakub and Lee, James and Quinn, James and Kiros, Jamie and Kalra, Janvi and Samaroo, Jasmyn and Wang, Jason and Wolfe, Jason and Chen, Jay and Wang, Jay and Harb, Jean and Han, Jeffrey and Wang, Jeffrey and Zhao, Jennifer and Chen, Jeremy and Yang, Jerene and Tworek, Jerry and Chand, Jesse and Landon, Jessica and Liang, Jessica and Lin, Ji and Liu, Jiancheng and Wang, Jianfeng and Tang, Jie and Yin, Jihan and Jang, Joanne and Morris, Joel and Flynn, Joey and Ferstad, Johannes and Heidecke, Johannes and Fishbein, John and Hallman, John and Grant, Jonah and Chien, Jonathan and Gordon, Jonathan and Park, Jongsoo and Liss, Jordan and Kraaijeveld, Jos and Guay, Joseph and Mo, Joseph and Lawson, Josh and {McGrath}, Josh and Vendrow, Joshua and Jiao, Joy and Lee, Julian and Steele, Julie and Wang, Julie and Mao, Junhua and Chen, Kai and Hayashi, Kai and Xiao, Kai and Salahi, Kamyar and Wu, Kan and Sekhri, Karan and Sharma, Karan and Singhal, Karan and Li, Karen and Nguyen, Kenny and Gu-Lemberg, Keren and King, Kevin and Liu, Kevin and Stone, Kevin and Yu, Kevin and Ying, Kristen and Georgiev, Kristian and Lim, Kristie and Tirumala, Kushal and Miller, Kyle and Ahmad, Lama and Lv, Larry and Clare, Laura and Fauconnet, Laurance and Itow, Lauren and Yang, Lauren and Romaniuk, Laurentia and Anise, Leah and Byron, Lee and Pathak, Leher and Maksin, Leon and Lo, Leyan and Ho, Leyton and Jing, Li and Wu, Liang and Xiong, Liang and Mamitsuka, Lien and Yang, Lin and {McCallum}, Lindsay and Held, Lindsey and Bourgeois, Liz and Engstrom, Logan and Kuhn, Lorenz and Feuvrier, Louis and Zhang, Lu and Switzer, Lucas and Kondraciuk, Lukas and Kaiser, Lukasz and Joglekar, Manas and Singh, Mandeep and Shah, Mandip and Stratta, Manuka and Williams, Marcus and Chen, Mark and Sun, Mark and Cayton, Marselus and Li, Martin and Zhang, Marvin and Aljubeh, Marwan and Nichols, Matt and Haines, Matthew and Schwarzer, Max and Gupta, Mayank and Shah, Meghan and Huang, Melody and Dong, Meng and Wang, Mengqing and Glaese, Mia and Carroll, Micah and Lampe, Michael and Malek, Michael and Sharman, Michael and Zhang, Michael and Wang, Michele and Pokrass, Michelle and Florian, Mihai and Pavlov, Mikhail and Wang, Miles and Chen, Ming and Wang, Mingxuan and Feng, Minnia and Bavarian, Mo and Lin, Molly and Abdool, Moose and Rohaninejad, Mostafa and Soto, Nacho and Staudacher, Natalie and {LaFontaine}, Natan and Marwell, Nathan and Liu, Nelson and Preston, Nick and Turley, Nick and Ansman, Nicklas and Blades, Nicole and Pancha, Nikil and Mikhaylin, Nikita and Felix, Niko and Handa, Nikunj and Rai, Nishant and Keskar, Nitish and Brown, Noam and Nachum, Ofir and Boiko, Oleg and Murk, Oleg and Watkins, Olivia and Gleeson, Oona and Mishkin, Pamela and Lesiewicz, Patryk and Baltescu, Paul and Belov, Pavel and Zhokhov, Peter and Pronin, Philip and Guo, Phillip and Thacker, Phoebe and Liu, Qi and Yuan, Qiming and Liu, Qinghua and Dias, Rachel and Puckett, Rachel and Arora, Rahul and Mullapudi, Ravi Teja and Gaon, Raz and Miyara, Reah and Song, Rennie and Aggarwal, Rishabh and Marsan, R. J. and Yemiru, Robel and Xiong, Robert and Kshirsagar, Rohan and Nuttall, Rohan and Tsiupa, Roman and Eldan, Ronen and Wang, Rose and James, Roshan and Ziv, Roy and Shu, Rui and Nigmatullin, Ruslan and Jain, Saachi and Talaie, Saam and Altman, Sam and Arnesen, Sam and Toizer, Sam and Toyer, Sam and Miserendino, Samuel and Agarwal, Sandhini and Yoo, Sarah and Heon, Savannah and Ethersmith, Scott and Grove, Sean and Taylor, Sean and Bubeck, Sebastien and Banesiu, Sever and Amdo, Shaokyi and Zhao, Shengjia and Wu, Sherwin and Santurkar, Shibani and Zhao, Shiyu and Chaudhuri, Shraman Ray and Krishnaswamy, Shreyas and Shuaiqi and Xia and Cheng, Shuyang and Anadkat, Shyamal and Fishman, Simón Posada and Tobin, Simon and Fu, Siyuan and Jain, Somay and Mei, Song and Egoian, Sonya and Kim, Spencer and Golden, Spug and Mah, S. Q. and Lin, Steph and Imm, Stephen and Sharpe, Steve and Yadlowsky, Steve and Choudhry, Sulman and Eum, Sungwon and Sanjeev, Suvansh and Khan, Tabarak and Stramer, Tal and Wang, Tao and Xin, Tao and Gogineni, Tarun and Christianson, Taya and Sanders, Ted and Patwardhan, Tejal and Degry, Thomas and Shadwell, Thomas and Fu, Tianfu and Gao, Tianshi and Garipov, Timur and Sriskandarajah, Tina and Sherbakov, Toki and Kaftan, Tomer and Hiratsuka, Tomo and Wang, Tongzhou and Song, Tony and Zhao, Tony and Peterson, Troy and Kharitonov, Val and Chernova, Victoria and Kosaraju, Vineet and Kuo, Vishal and Pong, Vitchyr and Verma, Vivek and Petrov, Vlad and Jiang, Wanning and Zhang, Weixing and Zhou, Wenda and Xie, Wenlei and Zhan, Wenting and {McCabe}, Wes and {DePue}, Will and Ellsworth, Will and Bain, Wulfie and Thompson, Wyatt and Chen, Xiangning and Qi, Xiangyu and Xiang, Xin and Shi, Xinwei and Dubois, Yann and Yu, Yaodong and Khakbaz, Yara and Wu, Yifan and Qian, Yilei and Lee, Yin Tat and Chen, Yinbo and Zhang, Yizhen and Xiong, Yizhong and Tian, Yonglong and Cha, Young and Bai, Yu and Yang, Yu and Yuan, Yuan and Li, Yuanzhi and Zhang, Yufeng and Yang, Yuguang and Jin, Yujia and Jiang, Yun and Wang, Yunyun and Wang, Yushi and Liu, Yutian and Stubenvoll, Zach and Dou, Zehao and Wu, Zheng and Wang, Zhigang},
	urldate = {2026-04-27},
	date = {2025-12-19},
	eprinttype = {arxiv},
	eprint = {2601.03267 [cs]},
}

@misc{hcompany_holo3_2026,
	title = {Holo3 - Open Foundation Models for Navigation and Computer Use Agents},
	url = {https://huggingface.co/Hcompany/Holo3-35B-A3B},
	author = {{HCompany}},
	date = {2026},
}

@misc{team_kimi_2026,
	title = {Kimi K2.5: Visual Agentic Intelligence},
	url = {http://arxiv.org/abs/2602.02276},
	doi = {10.48550/arXiv.2602.02276},
	shorttitle = {Kimi K2.5},
	number = {{arXiv}:2602.02276},
	publisher = {{arXiv}},
	author = {Team, Kimi and Bai, Tongtong and Bai, Yifan and Bao, Yiping and Cai, S. H. and Cao, Yuan and Charles, Y. and Che, H. S. and Chen, Cheng and Chen, Guanduo and Chen, Huarong and Chen, Jia and Chen, Jiahao and Chen, Jianlong and Chen, Jun and Chen, Kefan and Chen, Liang and Chen, Ruijue and Chen, Xinhao and Chen, Yanru and Chen, Yanxu and Chen, Yicun and Chen, Yimin and Chen, Yingjiang and Chen, Yuankun and Chen, Yujie and Chen, Yutian and Chen, Zhirong and Chen, Ziwei and Cheng, Dazhi and Chu, Minghan and Cui, Jialei and Deng, Jiaqi and Diao, Muxi and Ding, Hao and Dong, Mengfan and Dong, Mengnan and Dong, Yuxin and Dong, Yuhao and Du, Angang and Du, Chenzhuang and Du, Dikang and Du, Lingxiao and Du, Yulun and Fan, Yu and Fang, Shengjun and Feng, Qiulin and Feng, Yichen and Fu, Garimugai and Fu, Kelin and Gao, Hongcheng and Gao, Tong and Ge, Yuyao and Geng, Shangyi and Gong, Chengyang and Gong, Xiaochen and Gongque, Zhuoma and Gu, Qizheng and Gu, Xinran and Gu, Yicheng and Guan, Longyu and Guo, Yuanying and Hao, Xiaoru and He, Weiran and He, Wenyang and He, Yunjia and Hong, Chao and Hu, Hao and Hu, Jiaxi and Hu, Yangyang and Hu, Zhenxing and Huang, Ke and Huang, Ruiyuan and Huang, Weixiao and Huang, Zhiqi and Jiang, Tao and Jiang, Zhejun and Jin, Xinyi and Jing, Yu and Lai, Guokun and Li, Aidi and Li, C. and Li, Cheng and Li, Fang and Li, Guanghe and Li, Guanyu and Li, Haitao and Li, Haoyang and Li, Jia and Li, Jingwei and Li, Junxiong and Li, Lincan and Li, Mo and Li, Weihong and Li, Wentao and Li, Xinhang and Li, Xinhao and Li, Yang and Li, Yanhao and Li, Yiwei and Li, Yuxiao and Li, Zhaowei and Li, Zheming and Liao, Weilong and Lin, Jiawei and Lin, Xiaohan and Lin, Zhishan and Lin, Zichao and Liu, Cheng and Liu, Chenyu and Liu, Hongzhang and Liu, Liang and Liu, Shaowei and Liu, Shudong and Liu, Shuran and Liu, Tianwei and Liu, Tianyu and Liu, Weizhou and Liu, Xiangyan and Liu, Yangyang and Liu, Yanming and Liu, Yibo and Liu, Yuanxin and Liu, Yue and Liu, Zhengying and Liu, Zhongnuo and Lu, Enzhe and Lu, Haoyu and Lu, Zhiyuan and Luo, Junyu and Luo, Tongxu and Luo, Yashuo and Ma, Long and Ma, Yingwei and Mao, Shaoguang and Mei, Yuan and Men, Xin and Meng, Fanqing and Meng, Zhiyong and Miao, Yibo and Ni, Minqing and Ouyang, Kun and Pan, Siyuan and Pang, Bo and Qian, Yuchao and Qin, Ruoyu and Qin, Zeyu and Qiu, Jiezhong and Qu, Bowen and Shang, Zeyu and Shao, Youbo and Shen, Tianxiao and Shen, Zhennan and Shi, Juanfeng and Shi, Lidong and Shi, Shengyuan and Song, Feifan and Song, Pengwei and Song, Tianhui and Song, Xiaoxi and Su, Hongjin and Su, Jianlin and Su, Zhaochen and Sui, Lin and Sun, Jinsong and Sun, Junyao and Sun, Tongyu and Sung, Flood and Tai, Yunpeng and Tang, Chuning and Tang, Heyi and Tang, Xiaojuan and Tang, Zhengyang and Tao, Jiawen and Teng, Shiyuan and Tian, Chaoran and Tian, Pengfei and Wang, Ao and Wang, Bowen and Wang, Chensi and Wang, Chuang and Wang, Congcong and Wang, Dingkun and Wang, Dinglu and Wang, Dongliang and Wang, Feng and Wang, Hailong and Wang, Haiming and Wang, Hengzhi and Wang, Huaqing and Wang, Hui and Wang, Jiahao and Wang, Jinhong and Wang, Jiuzheng and Wang, Kaixin and Wang, Linian and Wang, Qibin and Wang, Shengjie and Wang, Shuyi and Wang, Si and Wang, Wei and Wang, Xiaochen and Wang, Xinyuan and Wang, Yao and Wang, Yejie and Wang, Yipu and Wang, Yiqin and Wang, Yucheng and Wang, Yuzhi and Wang, Zhaoji and Wang, Zhaowei and Wang, Zhengtao and Wang, Zhexu and Wang, Zihan and Wang, Zizhe and Wei, Chu and Wei, Ming and Wen, Chuan and Wen, Zichen and Wu, Chengjie and Wu, Haoning and Wu, Junyan and Wu, Rucong and Wu, Wenhao and Wu, Yuefeng and Wu, Yuhao and Wu, Yuxin and Wu, Zijian and Xiao, Chenjun and Xie, Jin and Xie, Xiaotong and Xie, Yuchong and Xin, Yifei and Xing, Bowei and Xu, Boyu and Xu, Jianfan and Xu, Jing and Xu, Jinjing and Xu, L. H. and Xu, Lin and Xu, Suting and Xu, Weixin and Xu, Xinbo and Xu, Xinran and Xu, Yangchuan and Xu, Yichang and Xu, Yuemeng and Xu, Zelai and Xu, Ziyao and Yan, Junjie and Yan, Yuzi and Yang, Guangyao and Yang, Hao and Yang, Junwei and Yang, Kai and Yang, Ningyuan and Yang, Ruihan and Yang, Xiaofei and Yang, Xinlong and Yang, Ying and Yang, Yi and Yang, Yi and Yang, Zhen and Yang, Zhilin and Yang, Zonghan and Yao, Haotian and Ye, Dan and Ye, Wenjie and Ye, Zhuorui and Yin, Bohong and Yu, Chengzhen and Yu, Longhui and Yu, Tao and Yu, Tianxiang and Yuan, Enming and Yuan, Mengjie and Yuan, Xiaokun and Yue, Yang and Zeng, Weihao and Zha, Dunyuan and Zhan, Haobing and Zhang, Dehao and Zhang, Hao and Zhang, Jin and Zhang, Puqi and Zhang, Qiao and Zhang, Rui and Zhang, Xiaobin and Zhang, Y. and Zhang, Yadong and Zhang, Yangkun and Zhang, Yichi and Zhang, Yizhi and Zhang, Yongting and Zhang, Yu and Zhang, Yushun and Zhang, Yutao and Zhang, Yutong and Zhang, Zheng and Zhao, Chenguang and Zhao, Feifan and Zhao, Jinxiang and Zhao, Shuai and Zhao, Xiangyu and Zhao, Yikai and Zhao, Zijia and Zheng, Huabin and Zheng, Ruihan and Zheng, Shaojie and Zheng, Tengyang and Zhong, Junfeng and Zhong, Longguang and Zhong, Weiming and Zhou, M. and Zhou, Runjie and Zhou, Xinyu and Zhou, Zaida and Zhu, Jinguo and Zhu, Liya and Zhu, Xinhao and Zhu, Yuxuan and Zhu, Zhen and Zhuang, Jingze and Zhuang, Weiyu and Zou, Ying and Zu, Xinxing},
	urldate = {2026-04-27},
	date = {2026-02-02},
	eprinttype = {arxiv},
	eprint = {2602.02276 [cs]},
}

@misc{qwen_qwen35_2026,
	title = {Qwen3.5: Accelerating Productivity with Native Multimodal Agents},
	url = {https://qwen.ai/blog?id=qwen3.5},
	author = {Qwen},
	date = {2026-02},
}

@misc{anthropic_claude_2025,
	title = {Claude 4},
	url = {https://www.anthropic.com},
	author = {{Anthropic}},
	date = {2025},
}

@inproceedings{yao_react_2023,
	title = {{ReAct}: Synergizing Reasoning and Acting in Language Models},
	url = {http://arxiv.org/abs/2210.03629},
	shorttitle = {{ReAct}},
	booktitle = {International Conference on Learning Representations ({ICLR})},
	author = {Yao, Shunyu and Zhao, Jeffrey and Yu, Dian and Du, Nan and Shafran, Izhak and Narasimhan, Karthik and Cao, Yuan},
	date = {2023},
	eprinttype = {arxiv},
	eprint = {2210.03629},
}

@inproceedings{yang_aria-ui_2025,
	location = {Vienna, Austria},
	title = {Aria-{UI}: Visual Grounding for {GUI} Instructions},
	url = {https://aclanthology.org/2025.findings-acl.1152/},
	doi = {10.18653/v1/2025.findings-acl.1152},
	shorttitle = {Aria-{UI}},
	pages = {22418--22433},
	booktitle = {Findings of the Association for Computational Linguistics: {ACL} 2025},
	publisher = {Association for Computational Linguistics},
	author = {Yang, Yuhao and Wang, Yue and Li, Dongxu and Luo, Ziyang and Chen, Bei and Huang, Chao and Li, Junnan},
	urldate = {2026-04-24},
	date = {2025-07},
}

@misc{wu_gui-actor_2025,
	title = {{GUI}-Actor: Coordinate-Free Visual Grounding for {GUI} Agents},
	url = {http://arxiv.org/abs/2506.03143},
	doi = {10.48550/arXiv.2506.03143},
	shorttitle = {{GUI}-Actor},
	publisher = {{arXiv}},
	author = {Wu, Qianhui and Cheng, Kanzhi and Yang, Rui and Zhang, Chaoyun and Yang, Jianwei and Jiang, Huiqiang and Mu, Jian and Peng, Baolin and Qiao, Bo and Tan, Reuben and Qin, Si and Liden, Lars and Lin, Qingwei and Zhang, Huan and Zhang, Tong and Zhang, Jianbing and Zhang, Dongmei and Gao, Jianfeng},
	urldate = {2026-04-24},
	date = {2025-06-03},
	eprinttype = {arxiv},
	eprint = {2506.03143 [cs]},
	note = {Issue: {arXiv}:2506.03143},
}

@misc{gou_navigating_2024,
	title = {Navigating the Digital World as Humans Do: Universal Visual Grounding for {GUI} Agents},
	url = {http://arxiv.org/abs/2410.05243},
	doi = {10.48550/arXiv.2410.05243},
	shorttitle = {{UGround}},
	publisher = {{arXiv}},
	author = {Gou, Boyu and Wang, Ruohan and Zheng, Boyuan and Xie, Yanan and Chang, Cheng and Shu, Yiheng and Sun, Huan and Su, Yu},
	urldate = {2026-04-24},
	date = {2024-10-07},
	eprinttype = {arxiv},
	eprint = {2410.05243 [cs]},
	note = {Issue: {arXiv}:2410.05243},
}

@misc{wu_os-atlas_2024,
	title = {{OS}-{ATLAS}: A Foundation Action Model for Generalist {GUI} Agents},
	url = {http://arxiv.org/abs/2410.23218},
	doi = {10.48550/arXiv.2410.23218},
	shorttitle = {{OS}-{ATLAS}},
	publisher = {{arXiv}},
	author = {Wu, Zhiyong and Wu, Zhenyu and Xu, Fangzhi and Wang, Yian and Sun, Qiushi and Jia, Chengyou and Cheng, Kanzhi and Ding, Zichen and Chen, Liheng and Liang, Paul Pu and Qiao, Yu},
	urldate = {2026-04-22},
	date = {2024-10-30},
	eprinttype = {arxiv},
	eprint = {2410.23218 [cs]},
	note = {Issue: {arXiv}:2410.23218},
}

@misc{datacite_data_2025,
	title = {Data Citation Corpus Data File},
	url = {https://doi.org/10.5281/zenodo.14897662},
	doi = {10.5281/zenodo.14897662},
	publisher = {{DataCite}},
	author = {{DataCite} and {Make Data Count}},
	urldate = {2026-04-27},
	date = {2025},
}

@misc{kingma_adam_2017,
	title = {Adam: A Method for Stochastic Optimization},
	url = {http://arxiv.org/abs/1412.6980},
	doi = {10.48550/arXiv.1412.6980},
	shorttitle = {Adam},
	number = {{arXiv}:1412.6980},
	publisher = {{arXiv}},
	author = {Kingma, Diederik P. and Ba, Jimmy},
	urldate = {2026-04-28},
	date = {2017-01-30},
	eprinttype = {arxiv},
	eprint = {1412.6980 [cs]},
}

@article{ratner_snorkel_2017,
	title = {Snorkel: Rapid Training Data Creation with Weak Supervision},
	volume = {11},
	url = {http://arxiv.org/abs/1711.10160},
	doi = {10.14778/3157794.3157797},
	shorttitle = {Snorkel},
	number = {3},
	journaltitle = {Proceedings of the {VLDB} Endowment},
	author = {Ratner, Alexander and Bach, Stephen H. and Ehrenberg, Henry and Fries, Jason and Wu, Sen and Ré, Christopher},
	date = {2017-11-28},
	eprinttype = {arxiv},
	eprint = {1711.10160 [cs]},
}

@inproceedings{tobin_domain_2017,
	title = {Domain randomization for transferring deep neural networks from simulation to the real world},
	url = {http://arxiv.org/abs/1703.06907},
	doi = {10.1109/IROS.2017.8202133},
	booktitle = {2017 {IEEE}/{RSJ} International Conference on Intelligent Robots and Systems ({IROS})},
	author = {Tobin, Josh and Fong, Rachel and Ray, Alex and Schneider, Jonas and Zaremba, Wojciech and Abbeel, Pieter},
	date = {2017-03-20},
	eprinttype = {arxiv},
	eprint = {1703.06907 [cs]},
}

@inproceedings{li_docbank_2020,
	title = {{DocBank}: A Benchmark Dataset for Document Layout Analysis},
	url = {http://arxiv.org/abs/2006.01038},
	shorttitle = {{DocBank}},
	booktitle = {Proceedings of the 28th International Conference on Computational Linguistics ({COLING})},
	author = {Li, Minghao and Xu, Yiheng and Cui, Lei and Huang, Shaohan and Wei, Furu and Li, Zhoujun and Zhou, Ming},
	date = {2020-12-01},
	eprinttype = {arxiv},
	eprint = {2006.01038 [cs]},
}

@inproceedings{kim_ocr-free_2022,
	title = {{OCR}-free Document Understanding Transformer},
	url = {http://arxiv.org/abs/2111.15664},
	shorttitle = {Donut},
	booktitle = {European Conference on Computer Vision ({ECCV})},
	author = {Kim, Geewook and Hong, Teakgyu and Yim, Moonbin and Nam, Jeongyeon and Park, Jinyoung and Yim, Jinyeong and Hwang, Wonseok and Yun, Sangdoo and Han, Dongyoon and Park, Seunghyun},
	date = {2022-10-04},
	eprinttype = {arxiv},
	eprint = {2111.15664 [cs]},
}
\bibliographystyle{icml2026}

\appendix
\onecolumn

\section{Drag Actions in OSWorld and AndroidWorld}
\label{sec:appendix-agent-drag-trials}

This section lists every task in which at least one drag is required to succeed, clustered by application domain.

\subsection{OSWorld (50 / 361 tasks, 13.9\%)}

\begin{itemize}[leftmargin=*,itemsep=1pt,topsep=2pt,parsep=0pt]
  \item GIMP (image editing) - 6 tasks
  \item LibreOffice Calc (spreadsheets) - 18 tasks
  \item LibreOffice Impress (slides) - 7 tasks
  \item LibreOffice Writer (word processing) - 9 tasks
  \item Multi-application workflows - 9 tasks
  \item VS Code - 1 task
\end{itemize}

\subsection{AndroidWorld (96 / 116 tasks, 82.8\%)}

AndroidWorld tasks are clustered by target application.

\begin{itemize}[leftmargin=*,itemsep=1pt,topsep=2pt,parsep=0pt]
  \item Notes and documents (\texttt{Markor}) - 11 tasks
  \item Calendars and to-dos (\texttt{SimpleCalendar}, \texttt{Tasks}) - 20 tasks
  \item Recipes and expenses (\texttt{Broccoli}, \texttt{ProExpense}) - 18 tasks
  \item System settings (brightness, Wi-Fi, Bluetooth) - 14 tasks
  \item Media apps (\texttt{RetroMusic}, \texttt{VLC}, \texttt{AudioRecorder}, \texttt{Camera}, \texttt{SimpleDrawPro}, \texttt{OpenTracks}) - 14 tasks
  \item Communication (\texttt{SimpleSms}, \texttt{Contacts}) - 7 tasks
  \item Browser-based tasks (\texttt{Chrome}) - 3 tasks
  \item Maps (\texttt{OsmAnd}) - 3 tasks
  \item Clock and file management (\texttt{Clock}, \texttt{Files}, \texttt{Gallery}, \texttt{Notes}) - 6 tasks
\end{itemize}

\section{Alternative Ground-Truth Target Regions for Resize and Rotation}
\label{sec:appendix-bbox-alternatives}

\begin{figure}[h]
  \centering
  \begin{subfigure}[t]{\linewidth}
    \centering
    \includegraphics[width=\linewidth]{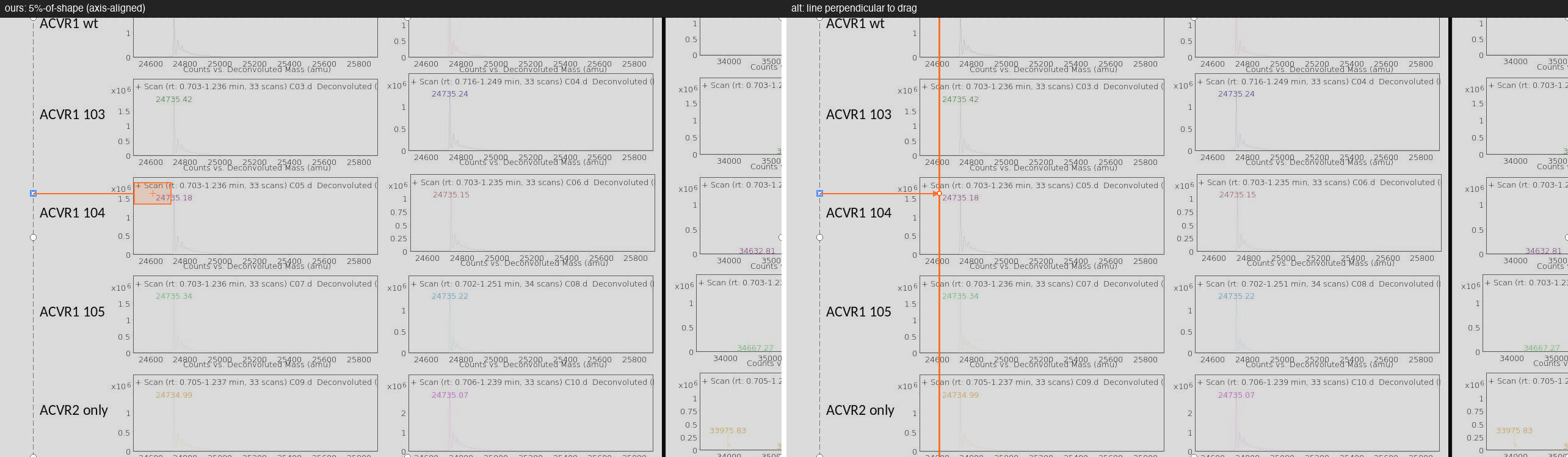}
    \caption{Resize along the horizontal axis. \textbf{Left:} our canonical target, a 5\%-of-shape axis-aligned box at the endpoint that matches the starting handle. \textbf{Right:} the alternative, a thin line perpendicular to the manipulated axis through the target, covering all drop points that yield the same resize outcome.}
    \label{fig:bbox-alternatives-resize}
  \end{subfigure}

  \vspace{0.6em}

  \begin{subfigure}[t]{\linewidth}
    \centering
    \includegraphics[width=\linewidth]{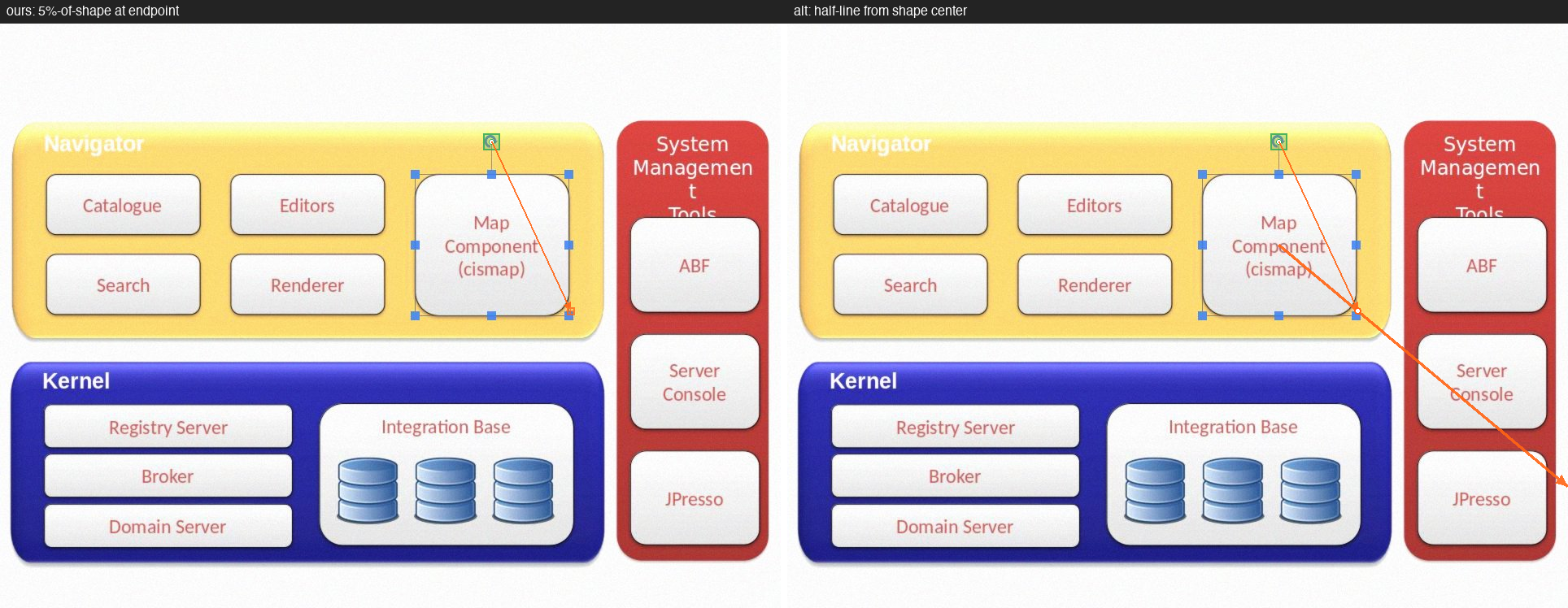}
    \caption{Rotation. \textbf{Left:} our canonical target, a 5\%-of-shape box at the endpoint on the circle centered at the shape's center. \textbf{Right:} the alternative, a half-line from the shape's center at the prescribed angle, covering all drop points that yield the same rotation.}
    \label{fig:bbox-alternatives-rotate}
  \end{subfigure}
  \caption{Canonical vs.\ alternative ground-truth target regions for actions with a continuum of valid drop points. We adopt the canonical convention (\emph{ours}, left in each pair) throughout the paper.}
  \label{fig:bbox-alternatives}
\end{figure}

As discussed in the main text, both resize and rotation actions admit a continuum of equally valid drop points. In our benchmark we use a \emph{canonical} ground-truth target - a small (5\% of shape) axis-aligned box at a single representative endpoint - for both training data generation and evaluation. An \emph{alternative} convention would be to make the ground-truth target cover the full set of equivalent drop points: for resize, a thin line orthogonal to the manipulated axis and passing through the target; for rotation, a half-line from the element's center at the prescribed angle. Figure~\ref{fig:bbox-alternatives} contrasts the two choices on a resize and a rotation example. We stick to the canonical convention for consistency with the training distribution, but both conventions are defensible.

\section{Qualitative End-to-End Impact}
\label{sec:qualitative-osworld}

\begin{figure*}[t]
  \centering
  \begin{subfigure}[b]{\linewidth}
    \centering
    \includegraphics[width=\linewidth]{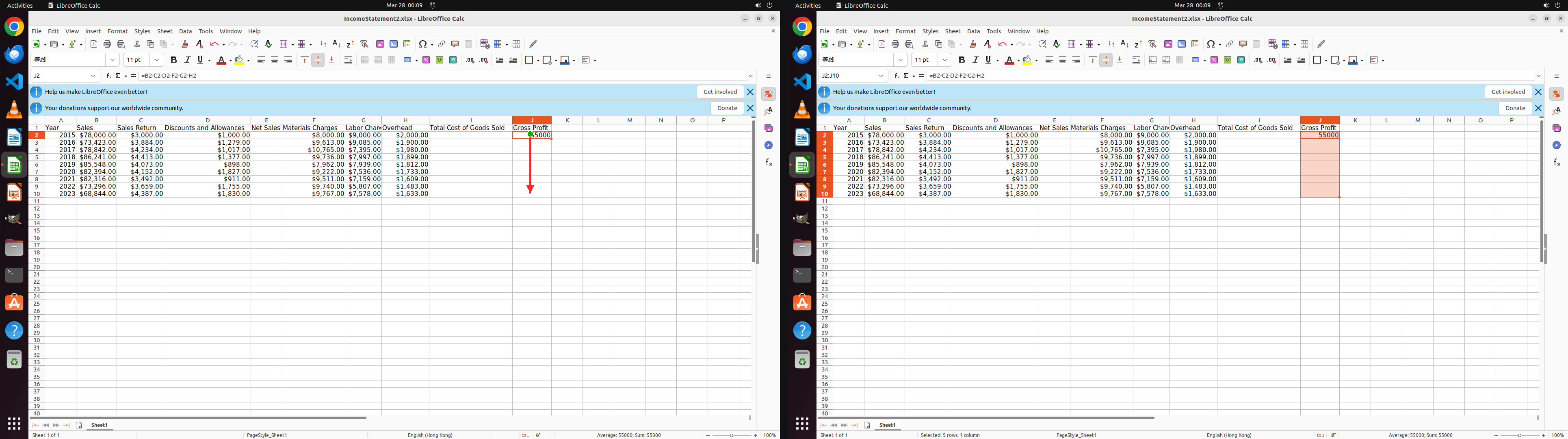}
    \caption{\textbf{Successful drag - computer-use-specialized policy (\texttt{holo3-35b-a3b}).} Left: observation before the drag, with the executed drag call overlaid in red. Right: next observation, in which the agent has produced a valid drag that selects the required range of cells; the trial ultimately succeeds.}
    \label{fig:qualitative-calc19-ok}
  \end{subfigure}

  \vspace{0.6em}

  \begin{subfigure}[b]{\linewidth}
    \centering
    \includegraphics[width=\linewidth]{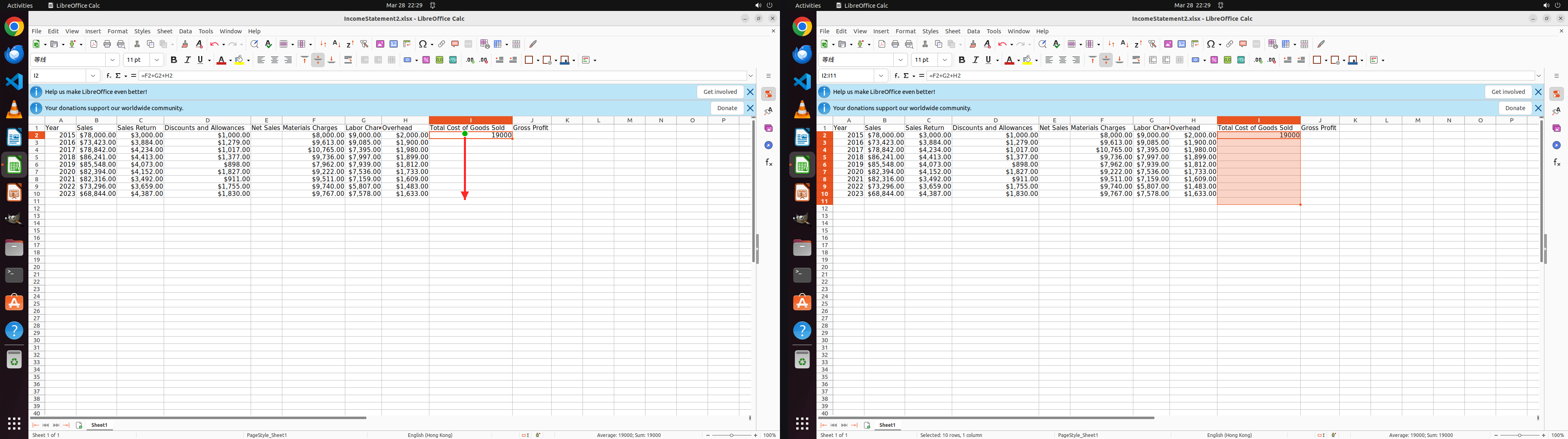}
    \caption{\textbf{Failed drag - generalist base policy (\texttt{qwen3-5-35b-a3b-fp8}).} Left: observation before the drag, with the executed drag call overlaid in red. Right: next observation, in which the agent 
    has not produced a valid drag, selecting the wrong range of cells; the trial ultimately fails.}
    \label{fig:qualitative-calc19-ko}
  \end{subfigure}
  \caption{Qualitative end-to-end comparison on the OSWorld task \texttt{libreoffice\_calc\_19}: the computer-use-specialized policy (\cref{fig:qualitative-calc19-ok}) executes a successful drag and solves the task, while the generalist base policy with the same architecture and parameter count (\cref{fig:qualitative-calc19-ko}) produces a failed drag on the same initial state; see \cref{sec:qualitative-osworld} for the full setup.}
  \label{fig:qualitative-osworld}
\end{figure*}

To illustrate how drag grounding translates to end-to-end computer-use performance,
Figure~\ref{fig:qualitative-osworld} contrasts two agents on the same OSWorld task
(\texttt{libreoffice\_calc\_19}): the task is to fill the \emph{Gross Profit} formula down cells
\texttt{J2:J10} of a spreadsheet. Both agents run the same ReAct-style
scaffold~\cite{yao_react_2023}: at each step, the agent policy observes a screenshot, reasons
about the current state in natural language, and emits an atomic action (click, drag, or
keystroke) that is executed in the environment; the resulting screenshot is then fed back into
the next step, until the task is completed or a step budget is exhausted. The scaffold, the
prompt, the tool set, the screen resolution, and the step budget are held fixed across the two
runs; the only difference between them is the underlying policy.

\section{Drag-and-Drop Intents: Examples}
\label{sec:appendix-intents}

This appendix shows example screenshots from our benchmark together with the
list of drag-and-drop intents collected for each one.

\begin{figure}[h]
  \centering
  \includegraphics[width=0.8\linewidth]{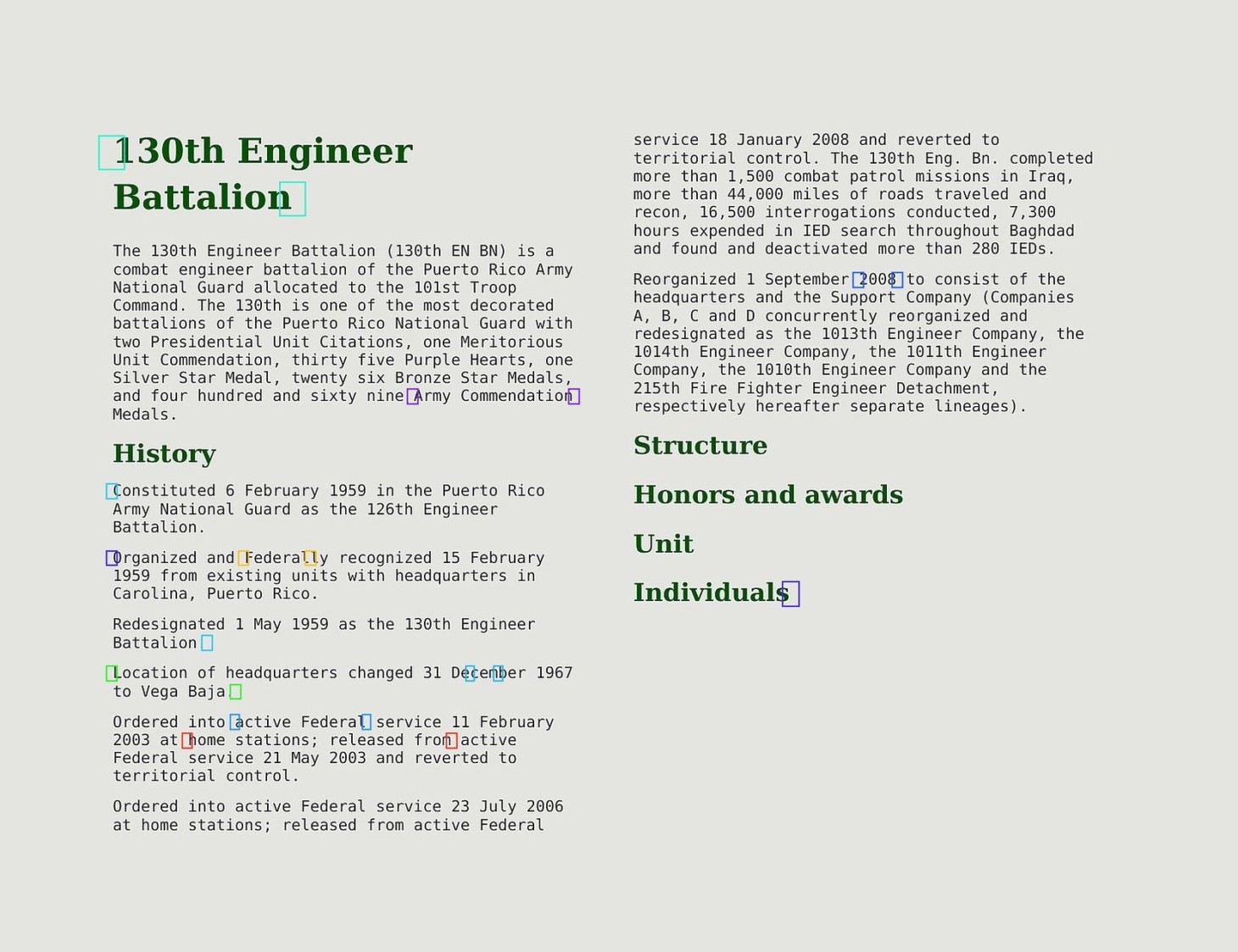}
  \caption{Text highlighting example.}
  \label{fig:intents-2065e6a0}
\end{figure}

Intents for \cref{fig:intents-2065e6a0}:
\begin{itemize}[itemsep=2pt,topsep=2pt,parsep=0pt]
  \item Highlight the range of the characters ``cem'' inside the word December
  \item Highlight the range of the text `active Federal'
  \item Drag across the institution: `Army Commendation Medals'
  \item Mark the extent of the paragraph starting with `Location of headquarters changed'
  \item Trace 3 paragraphs starting from paragraph 2
  \item Highlight the range of the title
  \item Mark the extent of the text home stations; released from
  \item Find the organization `Federal' and drag to highlight it in paragraph 3
  \item Click and drag to select the entire sentence starting with Organized and Federally recognized
  \item Swipe to select the word 2008 in the eighth paragraph
\end{itemize}

\begin{figure}[h]
  \centering
  \includegraphics[width=0.8\linewidth]{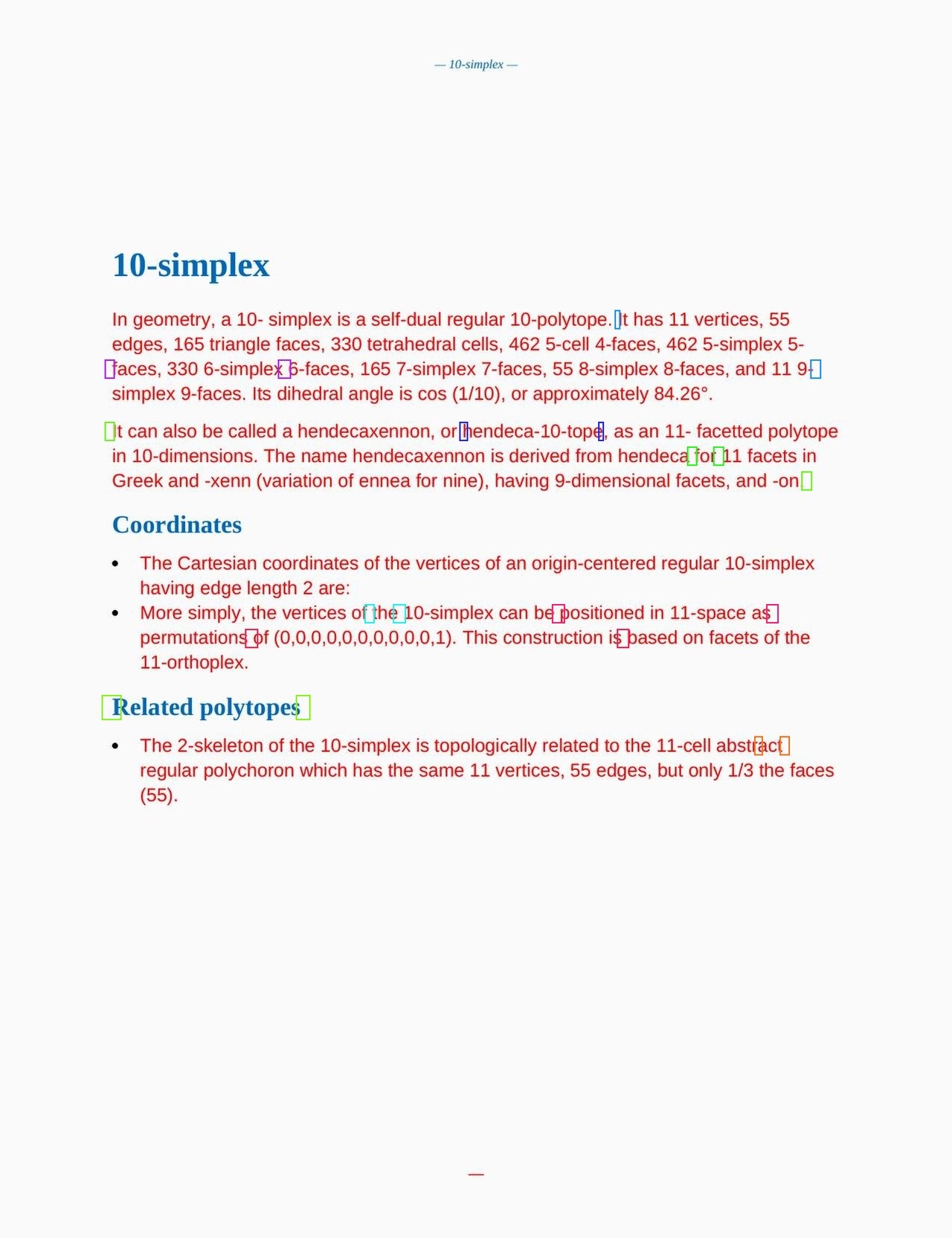}
  \caption{Text highlighting example.}
  \label{fig:intents-594335da}
\end{figure}

Intents for \cref{fig:intents-594335da}:
\begin{itemize}[itemsep=2pt,topsep=2pt,parsep=0pt]
  \item Mark the extent of the word `for' on the second line
  \item Please select from start to end of paragraph 2
  \item Select act within `abstract'
  \item Draw a selection over `positioned in 11-space as'
  \item Outline the date `hendeca-10-tope'
  \item Highlight the 3rd `the' on line 5
  \item Swipe to select the text `of (0,0,0,0,0,0,0,0,0,0,1). This construction is'
  \item Drag to highlight `faces, 330 6-simplex'
  \item Select the span of the sentence that ends with ``9-faces''
  \item Trace over the header `Related polytopes'
\end{itemize}

\begin{figure}[h]
  \centering
  \includegraphics[width=0.8\linewidth]{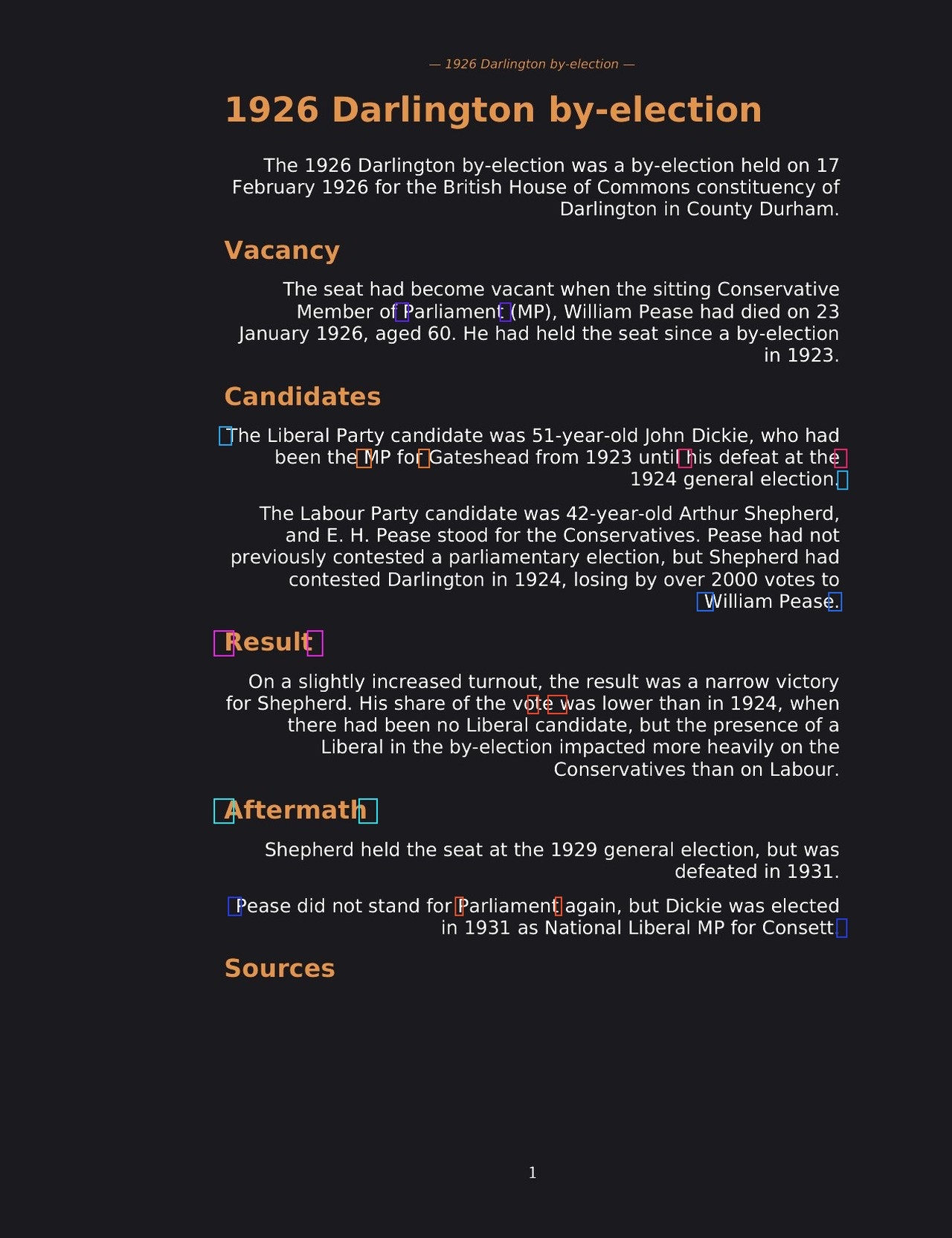}
  \caption{Text highlighting example.}
  \label{fig:intents-203fa28f}
\end{figure}

Intents for \cref{fig:intents-203fa28f}:
\begin{itemize}[itemsep=2pt,topsep=2pt,parsep=0pt]
  \item Outline the word ``Parliament'' in the seventh paragraph
  \item Drag across `MP for'
  \item Click and drag to select the paragraph that begins `The Liberal Party candidate'
  \item In paragraph 7, drag across the last sentence
  \item Select ``te'' within ``candidate'' in paragraph 5
  \item Trace `Parliament' in paragraph 2
  \item Highlight the range of the heading starting with `Result'
  \item Select the person's name: `William Pease' in paragraph 4
  \item Drag to select the text his defeat at the
\end{itemize}

\begin{figure}[h]
  \centering
  \includegraphics[width=0.8\linewidth]{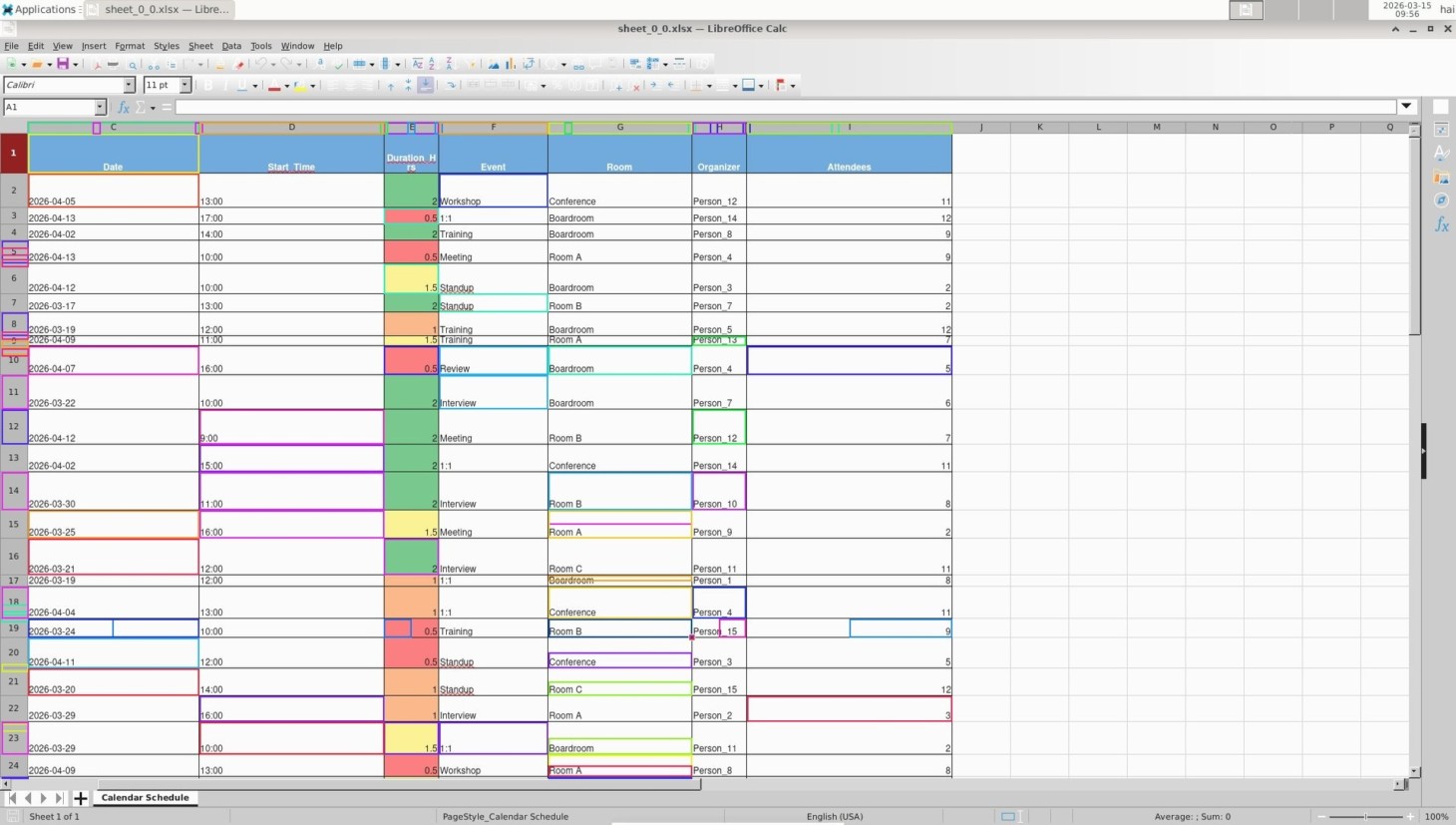}
  \caption{Cell selection example.}
  \label{fig:intents-96b9dcaa}
\end{figure}

Intents for \cref{fig:intents-96b9dcaa}:
\begin{itemize}[itemsep=2pt,topsep=2pt,parsep=0pt]
  \item Resize the 3rd column to be 0.5x its current width
  \item Resize the 8th row to be 1.75x its current height
  \item Drag to select from the cell containing `2026-04-07' to the cell containing `9:00'
  \item Choose from the `Duration\_Hrs' column to the `Room' column
  \item Use the fill handle at G19 to extend down by 4
  \item Resize row 5 to be 0.5x its current height
  \item Use the fill handle at G19 to extend down by 2
  \item Drag to extend to the `2026-03-24' cell to the left
  \item Drag to select from F7 to G10
  \item Drag from the Date column header to the Duration\_Hrs column header
  \item Select the range from `Review' to `2026-04-11'
  \item Mark columns `Start\_Time' to `Duration\_Hrs' across rows 22 to 23
  \item Make row 18 0.7x shorter
  \item Mark from the `Person\_10' cell in the Organizer column to the Start\_Time cell on that row
  \item Select from `Date' to `Room', rows 19--25
  \item Select the next 6 rows beginning at row 18
  \item Highlight 3 columns starting from column D
  \item Highlight cell E3 and the 3 cells below
  \item Mark the block spanning rows 15--16 and columns D--E
  \item Highlight the range from `2026-03-21' to `3'
  \item Select the range C21:D23
  \item Drag from `2026-04-05' to `Person\_10'
  \item Select 4 rows starting from row 11
  \item Drag from row 5 to row 8
  \item Choose columns `Event' to `Organizer' across rows 2 to 18
  \item Mark from the 3rd `Room A' to the 3rd `Conference' in the Room column
  \item Mark 14 rows starting from row 12
  \item Highlight cell H9 and the 3 cells below
  \item Extend up until row 15
  \item Use the fill handle at G19 to extend right by 1
  \item Resize row 9 to fit 10 characters of wrapped text
  \item Make the 6th column 0.4x narrower
  \item Select from `Start\_Time' to `Event', rows 22--23
  \item Drag the fill handle 2 cells left
  \item Highlight from the cell containing `Review' to the cell containing `2026-03-25'
  \item Resize the `Start\_Time' column to be 2x its current width
  \item Highlight from the cell containing `15:00' to the cell containing `2026-03-20'
  \item Make row 20 3x taller
  \item Choose from C1 to G24
  \item Extend down until row 24
  \item Drag the fill handle 2 cells up
  \item Highlight cells F11 through G14
  \item Select from the 1st occurrence of `5' in the Attendees column to the Duration\_Hrs cell on that row
  \item Extend the Room fill handle 1 row below
  \item Drag the fill handle right to column I
  \item Make the 1st column 0.4x narrower
  \item Drag to select the next 4 columns beginning at column F
  \item Widen the `Room' column by 2x
  \item Select columns E to G
  \item Drag from column E to column H
\end{itemize}

\begin{figure}[h]
  \centering
  \includegraphics[width=0.8\linewidth]{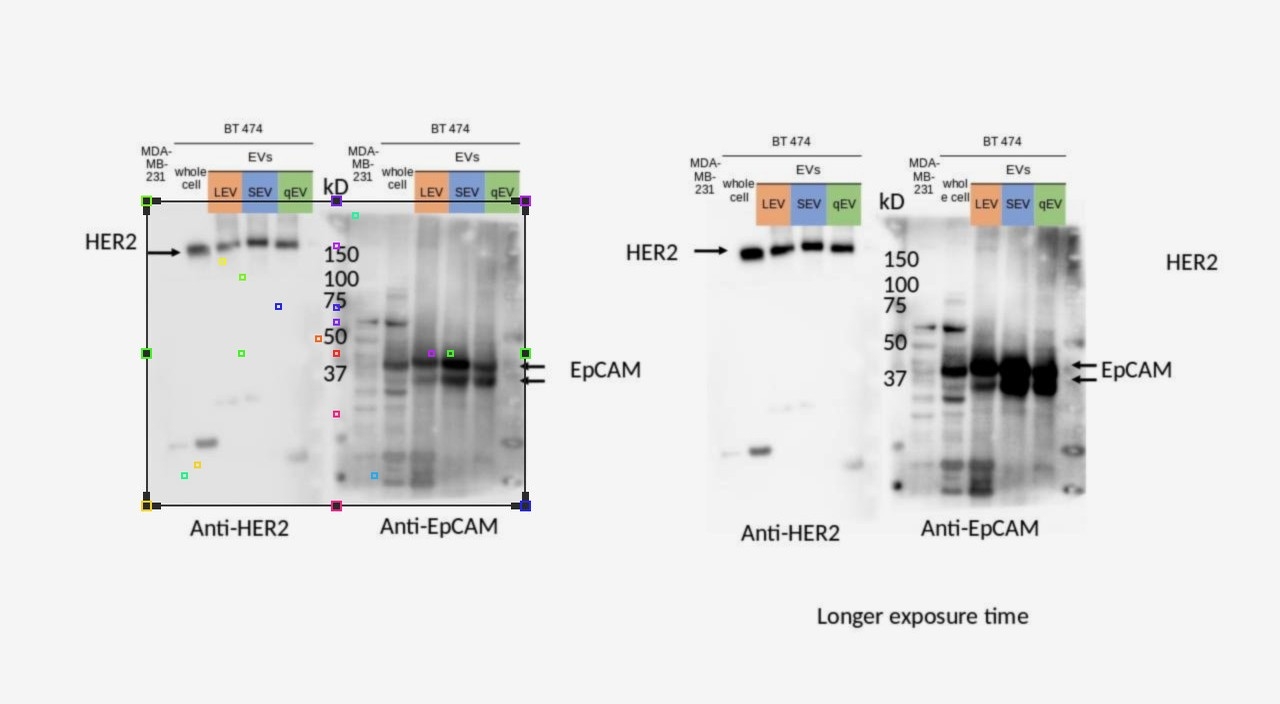}
  \caption{Element resizing example (crop).}
  \label{fig:intents-fd233d36}
\end{figure}

Intents for \cref{fig:intents-fd233d36}:
\begin{itemize}[itemsep=2pt,topsep=2pt,parsep=0pt]
  \item drag the bottom-right crop corner to crop the visible width by 40\% and cut height by 10\%
  \item trim the visible area by 25\% using the bottom-left crop corner
  \item trim the visible area by 50\% by dragging the right crop handle left
  \item crop the visible area by 15\% by dragging the top crop handle down
  \item pull the right crop handle left to trim the width to 80\%
  \item pull the bottom-left crop corner to crop the visible area by 19\%
  \item trim the visible area by 30\% using the bottom crop handle
  \item drag the top crop handle down to cut the area by 35\%
  \item cut the visible width by 25\% via the left crop handle
  \item crop the visible area proportionally to 30\% using the top-left crop corner
  \item trim the visible area to 56\% by dragging the top-left crop corner
  \item drag the bottom-right crop corner to cut width to 25\% and cut height to 50\%
  \item cut the area to 50\% by pulling the top crop handle down
  \item drag the left crop mark right to reach 75\% visible area
  \item cut the area by 75\% by pulling the right crop handle left
  \item trim the visible area proportionally by 88\% using the bottom-right crop corner
  \item trim the area proportionally to 64\% by dragging the top-left crop corner
  \item use the top-right crop handle to set visible width to 75\% and height to 50\%
  \item use the top-right crop handle to set visible width to 55\% and height to 95\%
  \item trim the area by 60\% via the bottom crop handle up
\end{itemize}

\begin{figure}[h]
  \centering
  \includegraphics[width=0.8\linewidth]{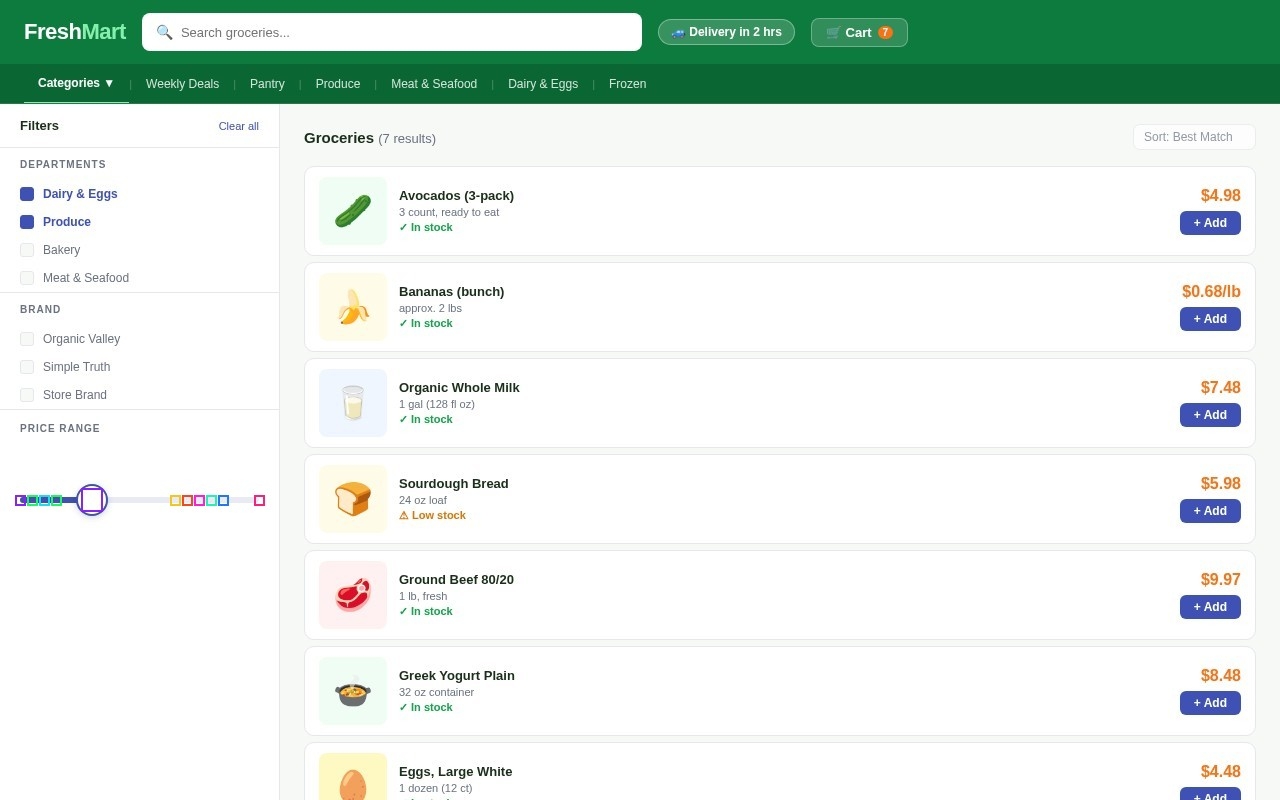}
  \caption{Slider manipulation example.}
  \label{fig:intents-b3e082f5}
\end{figure}

Intents for \cref{fig:intents-b3e082f5}:
\begin{itemize}[itemsep=2pt,topsep=2pt,parsep=0pt]
  \item Drag to the minimum value
  \item Knowing the price max is \$1000, drag to USD 800
  \item The slider runs 0 to \$1000; set the price to 700 dollars
  \item On a 0 to \$1000 scale, set the price to 50 dollars
  \item The slider runs 0 to \$1000; set the price to 850 dollars
  \item On a 0 to \$1000 scale, set the price to 100 dollars
  \item The price scale goes up to \$1000; set it to 150 dollars
  \item Put the slider at the three-quarter point
  \item Set the price to USD 650 (the max is \$1000)
  \item Set to \$1000 (the maximum)
\end{itemize}

\begin{figure}[h]
  \centering
  \includegraphics[width=0.8\linewidth]{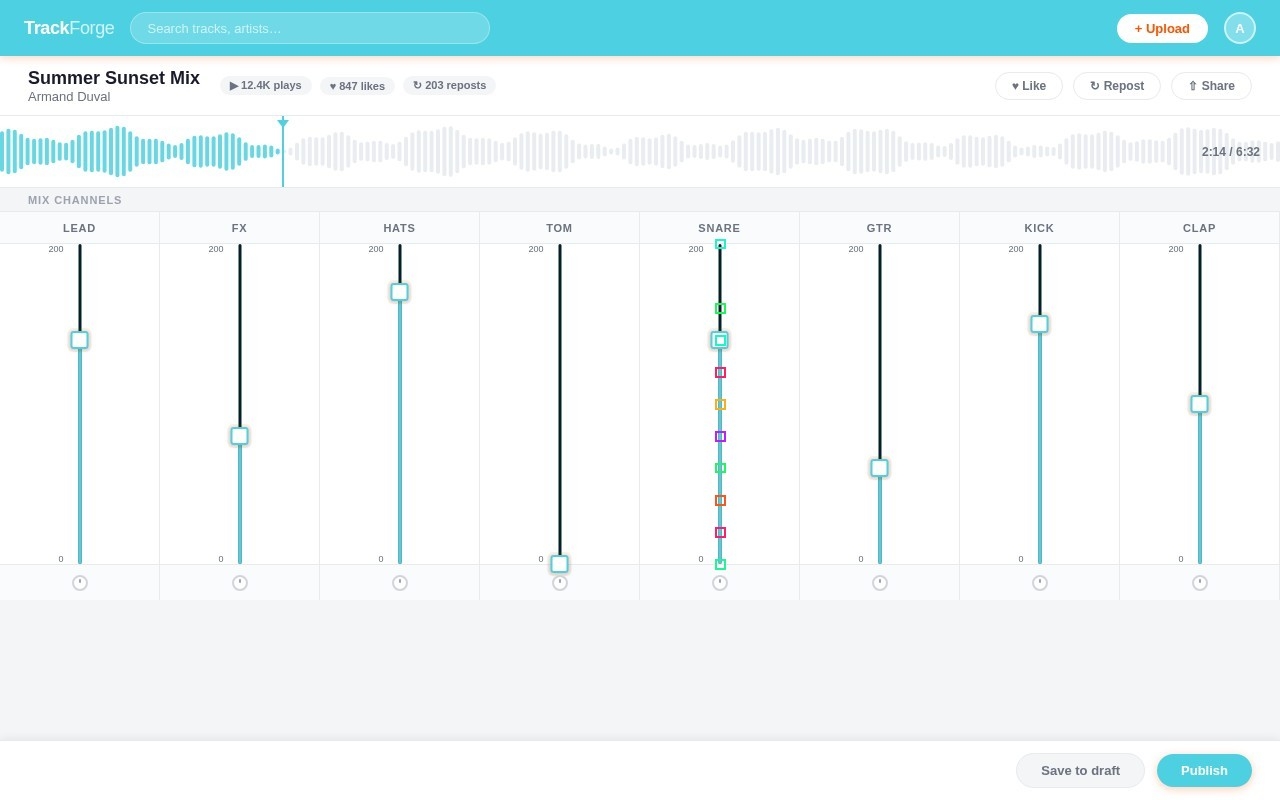}
  \caption{Slider manipulation example (vertical mixer fader).}
  \label{fig:intents-c2016c48}
\end{figure}

Intents for \cref{fig:intents-c2016c48}:
\begin{itemize}[itemsep=2pt,topsep=2pt,parsep=0pt]
  \item On SNARE: drag to the highest position
  \item On SNARE: set the volume to 100
  \item On SNARE: move the volume to 80
  \item On SNARE: put the volume at 40
  \item On SNARE: set to 0 (the minimum)
  \item On SNARE: change the volume to 60
  \item On SNARE: the volume ranges from 0 to 200; put it at 160
  \item On SNARE: change the volume to 20
  \item On SNARE: the slider runs 0 to 200; set the volume to 120
\end{itemize}

\end{document}